\documentclass[11pt]{article}

\usepackage[final]{acl}

\usepackage{times}
\usepackage{latexsym}

\usepackage[T1]{fontenc}

\usepackage[utf8]{inputenc}

\usepackage{microtype}

\usepackage{inconsolata}

\usepackage{graphicx}

\usepackage{amssymb}
\usepackage{amsmath}
\DeclareMathOperator*{\argmax}{argmax}

\DeclareMathOperator{\Norm}{Norm}
\usepackage{bbm}
\usepackage{booktabs}
\usepackage[normalem]{ulem}
\usepackage{subcaption}

\title{Repurposing Adversarial Perturbations for Continual Learning:\\ From Defense to Active Alignment}

\author{
 \textbf{Ran Liu\textsuperscript{1,2}},
 \textbf{Min Yu\textsuperscript{1,2}\thanks{Corresponding author.}},
 \textbf{Mingqi Liu\textsuperscript{1}},
 \textbf{Jianguo Jiang\textsuperscript{1,2}},
\\
 \textbf{Gang Li\textsuperscript{3}},
 \textbf{Rongsheng Li\textsuperscript{4}},
 \textbf{Ning Li\textsuperscript{1}},
 \textbf{Zhen Xu\textsuperscript{1,2}},
 \textbf{Weiqing Huang\textsuperscript{1,2}},
 \textbf{Ming Liu\textsuperscript{3}\thanks{Corresponding author.}}
\\
 \textsuperscript{1}Institute of Information Engineering, Chinese Academy of Sciences
\\
 \textsuperscript{2}School of Cyber Security, University of Chinese Academy of Sciences
\\
 \textsuperscript{3}Deakin University
 \quad
 \textsuperscript{4}Harbin Engineering University
\\
 \texttt{liuran@iie.ac.cn, yumin@iie.ac.cn, m.liu@deakin.edu.au}
}

\begin{document}
\maketitle
\begin{abstract}
In dynamic environments, large language models need to keep adapting to new tasks, but continual learning often suffers from forgetting, limited transfer, and vulnerability to adversarial perturbations. To address this, we present AdvCL, which repurposes adversarial perturbations as a geometric control signal for stable continual adaptation. AdvCL combines three plug-in modules: Intra-Smooth promotes local smoothness via small adversarial perturbations; Proto-Clip uses similarity clipping to prevent excessive alignment to current task prototype; and Inter-Align applies directional alignment toward previous task prototype to reduce representational gaps. Experiments show consistent gains in both standard performance and robustness, with lower forgetting and stronger transfer. We further analyze key mechanisms by quantifying the sensitivity of Intra-Smooth to perturbation settings and the effect of Inter-Align on task similarity and geometric distance. In summary, the modules provide complementary gains when combined, and each can also be integrated individually into diverse CL paradigms, including replay, regularization, and dynamic architectures, thereby offering a geometric control mechanism for continual learning.
\end{abstract}

\begin{figure}[t]
	\centering
	\includegraphics[width=0.99\linewidth]{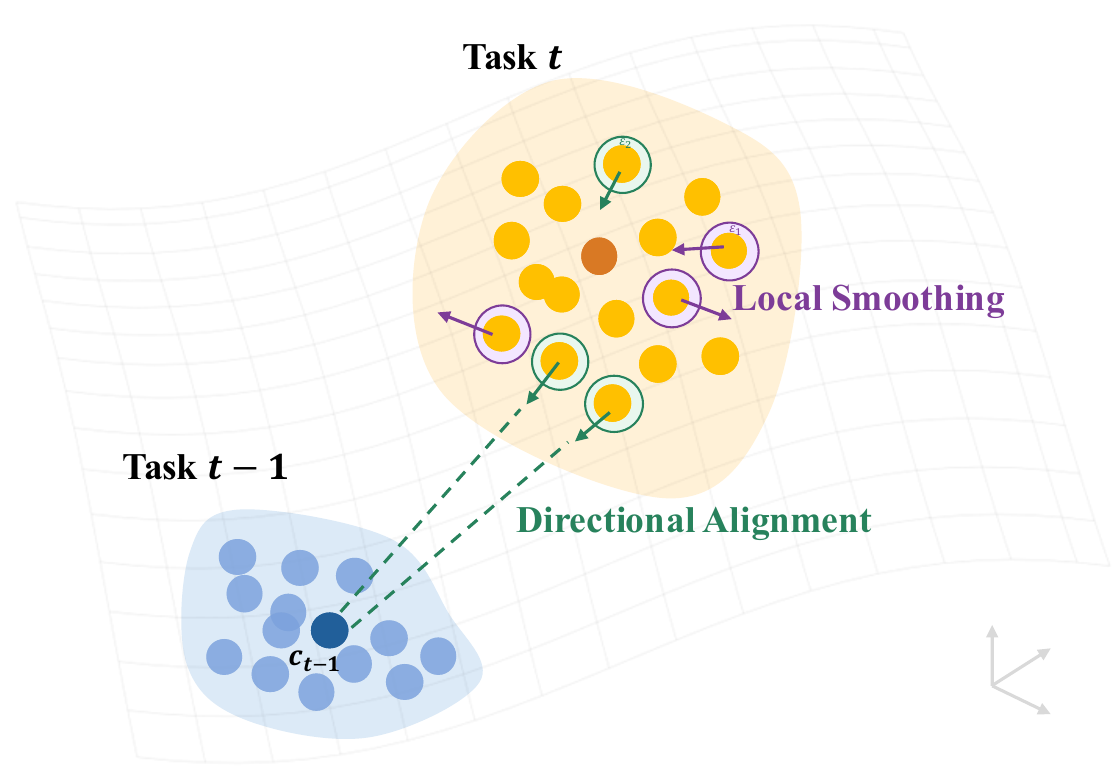}
	\caption{Illustration of local smoothing and directional alignment in representation space.
}
	\label{fig:geometry}
\end{figure}

\begin{figure*}[t]
	\centering
	\includegraphics[width=0.98\linewidth]{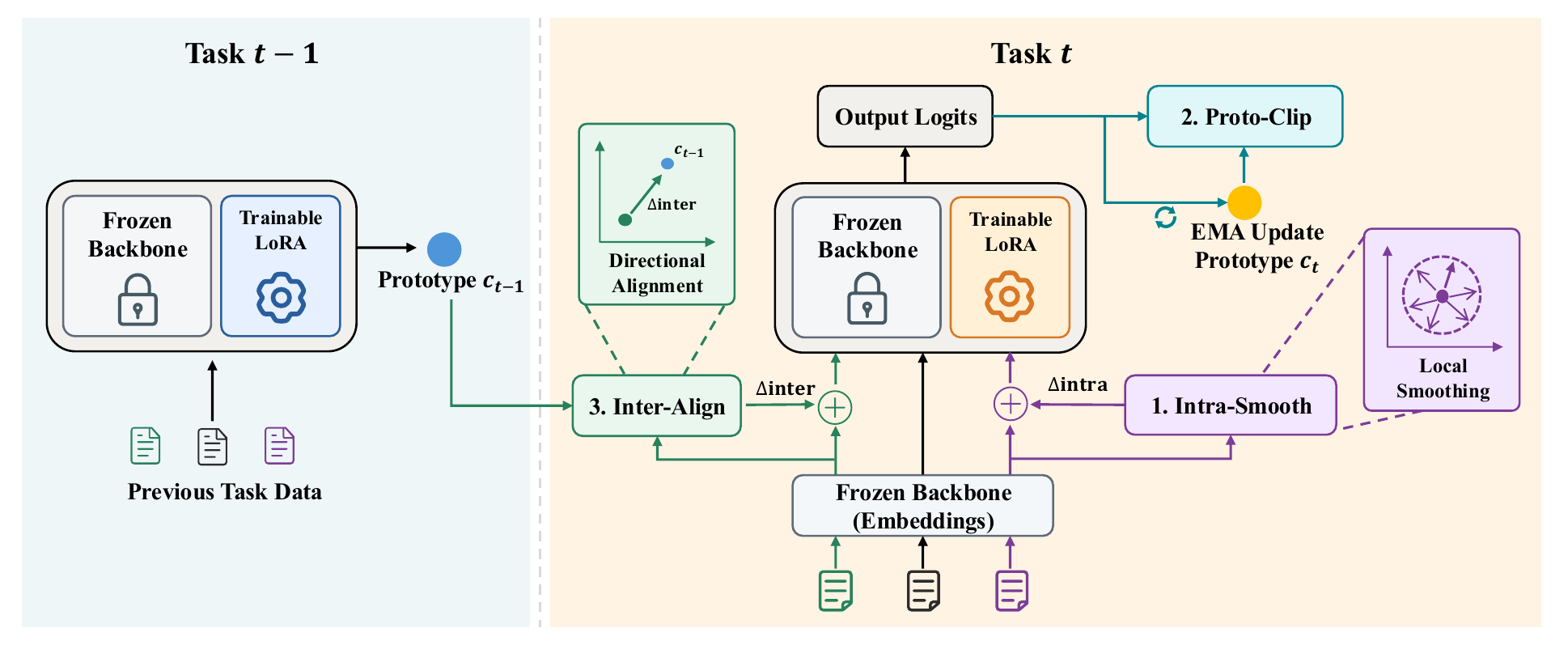}
	\caption{Overview of AdvCL. It integrates Intra-Smooth, Proto-Clip, and Inter-Align, and maintains a prototype for each task as a semantic anchor for similarity clipping and subsequent directional alignment.}
	\label{fig:framework}
\end{figure*}

\section{Introduction}
In dynamic environments, large language models (LLMs) must continually adapt to new tasks while preserving prior capabilities. Parameter efficient fine tuning (PEFT) supports this setting by updating only a small set of parameters \cite{lester-etal-2021-power,hu2022lora}, and is widely used in continual learning (CL). However, alignment does not ensure stable later adaptation. Models can exhibit rebound, where further fine tuning pulls behavior back toward the pretraining distribution \cite{ji-etal-2025-language-models}. This instability aligns with an optimization geometry view, since sharp minima are more sensitive to perturbations and tend to generalize unstably \cite{keskar2017on}, thereby amplifying drift and increasing vulnerability to noise and adversarial inputs. These issues can be traced to unstable representations and insufficiently smooth local geometry, which often manifest as forgetting, rebound, and brittleness.

This motivates a key question: can adversarial perturbations be repurposed from defense into a geometric control signal for stable transfer in CL? The intuition is that robustness reflects local geometric stability, which is tied to mitigating forgetting. Translating this intuition into a practical method is challenging. First, adversarial training involves a trade-off between neighborhood stability and standard accuracy \cite{pmlr-v97-zhang19p}, and continual updates can disrupt this stability so that robustness degrades over time \cite{ru2024maintainingadversarialrobustnesscontinuous}. Second, task similarity strongly influences transfer and forgetting \cite{NEURIPS2024_05cdc7fe}. Prior work characterizes task relationships to predict transfer feasibility \cite{Achille_2019_ICCV}. But for low similarity task pairs, mainstream methods mainly constrain drift to preserve old representations \cite{Rebuffi_2017_CVPR,Cha_2021_ICCV} instead of actively narrowing cross-task distances.

We argue that adversarial perturbations are valuable because their direction is controllable. As illustrated in Figure \ref{fig:geometry}, they support local smoothing to reduce geometric brittleness \cite{8417973,foret2021sharpnessaware} and directional alignment to actively shrink geometric distances between low similarity tasks, thereby improving transfer.

To this end, we present AdvCL, a multi-level adversarial perturbation framework for CL, shown in Figure \ref{fig:framework}. AdvCL maintains a prototype for the previous task and updates the current prototype with an exponential moving average. It includes three modules: (1) Intra-Smooth, which promotes local smoothing via small perturbations in a frozen embedding neighborhood; (2) Proto-Clip, which clips the similarity between current batch representations and current task prototype to prevent over alignment toward prototype; and (3) Inter-Align, which performs directional alignment by perturbing features toward the previous prototype, actively shrinking geometric distance between tasks.

Experiments show that AdvCL improves both standard performance and robustness under multiple evaluations, with lower forgetting and stronger transfer. We also find strong transferability of three modules, with complementary gains when combined. Finally, we analyze key mechanisms, including the sensitivity of Intra-Smooth to perturbation solvers and the effect of Inter-Align on task similarity and geometric distance.

Our main contributions are as follows:

(1) We introduce a new CL perspective that treats adversarial perturbations as a tool for local smoothing and cross-task directional alignment, providing a geometric route to improve stability and reduce forgetting. To our knowledge, this is the first to do so in PEFT-CL.

(2) We present AdvCL, a three-module framework that leverages neighborhood perturbations for local smoothing, similarity clipping to avoid over-alignment to the current prototype, and directional alignment for retention and cross-task stability.

(3) AdvCL jointly improves standard performance and robustness, yielding lower forgetting and stronger knowledge transfer. Moreover, modules are applicable to representative CL paradigms.

\section{Related Work}
\textbf{Continual learning} (CL) mitigates catastrophic forgetting via replay \cite{Rebuffi_2017_CVPR,huang-etal-2024-mitigating}, regularization \cite{doi:10.1073/pnas.1611835114,NEURIPS2024_ef626147}, and dynamic architectures \cite{rusu2022progressiveneuralnetworks,wang-etal-2024-rehearsal}. Parameter efficient fine tuning (PEFT) reduces training cost by updating only a small set of parameters \cite{pmlr-v97-houlsby19a,hu2022lora}. Recent PEFT-CL methods largely follow two directions: constraining PEFT updates in a subspace to reduce cross-task conflicts \cite{wang-etal-2023-orthogonal,lu-etal-2025-controlled}, or maintaining a module pool and composing modules via selection or routing to balance stability and plasticity \cite{zhao-etal-2024-sapt,han-etal-2025-slim}. In contrast, we enforce local smoothness via representation space perturbations, making PEFT updates more robust to perturbations and task distribution shifts, thus stabilizing continual adaptation.

\textbf{Representation space perturbations} have been studied for local smoothing and geometric regularization, for example through adversarial training that searches for worst-case directions \cite{Zhu2020FreeLB,jiang-etal-2020-smart} or through noise injection during forward pass \cite{aghajanyan2021better,jain2024neftune}. These methods mainly target single task generalization or offline adaptation, not cross-task consistency in CL. They often allow trainable embeddings or full parameter updates, which can drift semantic reference frame so that perturbations across tasks are no longer defined under the same reference frame, amplifying semantic anchor shifts and forgetting. In contrast, we fix embeddings and backbone representations, and use perturbations to enforce local smoothness and improve drift resistance of PEFT subspace updates rather than primarily targeting input robustness.

\textbf{Task similarity} strongly affects transfer and forgetting in CL. Prior work exploits task relations, for example by using task embeddings to measure similarity \cite{Achille_2019_ICCV} and by selecting task orders \cite{bell2022effecttaskorderingcontinual}. Another line stabilizes representations across tasks by anchoring prior knowledge with prototypes, representation alignment, or consistency learning \cite{cui-etal-2021-refining,wang-etal-2019-sentence,zhao-etal-2022-consistent}. Instead of treating similarity as a fixed signal or using alignment only to preserve old representations, we connect the two. Our active alignment paradigm uses similarity to trigger directional alignment on current data toward prior semantic anchors, shrinking representational gaps for low similarity task pairs and improving retention and backward transfer.

\section{AdvCL Framework}
AdvCL (as illustrated in Figure~\ref{fig:framework}) repurposes adversarial perturbations and auxiliary constraints as training signals, where Intra-Smooth enforces intra-task neighborhood consistency to promote local smoothing, Proto-Clip uses a prototype constraint to prevent over-alignment to current task, and Inter-Align encourages inter-task directional alignment to enhance retention and stability.

\subsection{Definitions}
We consider online continual learning over a task sequence $\{\mathcal{T}_t\}_{t=1}^{N}$, where task $t$ has dataset $D_t=\{(x,y)\}$. A single model is trained sequentially across tasks. At each task, only trainable parameters $\theta$ in parameter-efficient fine-tuning (PEFT) modules are updated while backbone parameters remain frozen. We write the model with trainable parameters $\theta$ as $f_\theta$, and denote its intermediate representations by $H_\theta(\cdot)$.

Given an input text $x=(w_1,\ldots,w_L)$, let $E(x)\in\mathbb{R}^{L\times d_e}$ be its embedding matrix, where $d_e$ is the embedding dimension. We use last layer token representations $H_\theta(x)\in\mathbb{R}^{L\times d}$ to form a text representation via mean pooling:
\[
h_\theta(x)=\mathrm{MeanPool}\big(H_\theta(x)\big)\in\mathbb{R}^{d}.
\]

We measure similarity using cosine similarity, defined as:
\[
S(\mathbf{u},\mathbf{v})
=
\frac{\mathbf{u}^\top\mathbf{v}}
{\|\mathbf{u}\|_2\,\|\mathbf{v}\|_2}.
\]

For embedding level perturbations, $\Delta\in\mathbb{R}^{L\times d_e}$ denotes a continuous perturbation added to $E(x)$, and $\nabla_{\Delta}$ denotes the gradient with respect to $\Delta$. Analysis of discrete recovery from continuous perturbations is provided in Appendix~\ref{app:discovery}.

\subsection{Intra-Task Local Smoothing (Intra-Smooth)}
Intra-Smooth promotes local smoothing by training on worst-case embedding perturbations via an inner–outer procedure, where the inner step approximates $\Delta^\star$ under budget $\varepsilon_{\mathrm{intra}}$, and the outer step updates trainable parameters $\theta$ on perturbed samples. From the current batch $\mathcal{B}$, we sample a subset $\mathcal{B}_{\mathrm{intra}}$. For any $(x,y)\in\mathcal{B}_{\mathrm{intra}}$, the inner perturbation is:
\begin{equation}
\label{eq:intra_inner}
\Delta_{\mathrm{intra}}^{\star}(x)\!=\!
\argmax_{\|\Delta\|_2\le \varepsilon_{\mathrm{intra}}}\!
\mathcal{L}_{\mathrm{sup}}\big(f_\theta(E(x)+\Delta),y\big),
\end{equation}
where $\mathcal{L}_{\mathrm{sup}}$ denotes supervised cross entropy loss. The inner maximization seeks an adversarial perturbation within the budget that maximizes the loss. We approximate Equation~\ref{eq:intra_inner} with $K_{\mathrm{intra}}$ steps of PGD \cite{madry2018towards}. Let $\Delta^{(0)}=0$ and iteratively update $\Delta^{(k)}$ by:
\begin{equation}
\label{eq:intra_pgd}
\begin{aligned}
\Delta^{(k)}
&=
\Pi_{\varepsilon_{\mathrm{intra}}}\!\big(
\Delta^{(k-1)}+\alpha_{\mathrm{intra}}\cdot
\Norm\!\big( \\
&\quad \nabla_{\Delta}\mathcal{L}_{\mathrm{sup}}
\big(f_\theta(E(x)+\Delta^{(k-1)}),y\big)
\big)
\big),
\end{aligned}
\end{equation}
where $\alpha_{\mathrm{intra}}=\varepsilon_{\mathrm{intra}}/K_{\mathrm{intra}}$, $\Norm(\cdot)$ is $\ell_2$ normalization, and $\Pi_{\varepsilon}$ projects onto the $\ell_2$ ball.

The outer adversarial loss is:
\begin{equation}
\label{eq:intra_outer}
\mathcal{L}_{\mathrm{intra}}
=
\mathbb{E}\!\left[
\mathcal{L}_{\mathrm{sup}}\!\big(f_\theta(E(x)+\Delta_{\mathrm{intra}}^{\star}(x)),y\big)
\right].
\end{equation}

Beyond the PGD-based solver, we also consider alternative inner solvers for generating $\Delta_{\mathrm{intra}}^{\star}(x)$.

\textbf{FGSM} computes a single update at $\Delta=0$ \cite{goodfellow2015explainingharnessingadversarialexamples}:
\begin{equation}
\label{eq:intra_fgsm}
\begin{aligned}
\Delta_{\mathrm{intra}}^{\star}(x)
&=
\varepsilon_{\mathrm{intra}}\cdot
\mathrm{sign}\big( \\
&\quad \nabla_{\Delta}\mathcal{L}_{\mathrm{sup}}\!\big(f_\theta(E(x)+\Delta),y\big)
\big)\big|_{\Delta=0} ,
\end{aligned}
\end{equation}
where $\mathrm{sign}(\cdot)$ denotes element-wise sign function. Unlike Equation~\ref{eq:intra_pgd}, FGSM uses an $\ell_\infty$ budget, applies one update, and does not involve projection.

\textbf{Rand} samples a random direction and scales it to $\ell_2$ budget:
\begin{equation}
\label{eq:intra_randl2}
\Delta_{\mathrm{intra}}^{\star}(x)
=\varepsilon_{\mathrm{intra}}\cdot \frac{\xi}{\|\xi\|_2},
\,\,
\xi\sim\mathcal{N}(0,I).
\end{equation}

\textbf{TRADES} replaces supervised loss with a KL consistency objective \cite{pmlr-v97-zhang19p}. The inner perturbation is:
\begin{equation}
\label{eq:intra_trades_inner}
\Delta_{\mathrm{intra}}^{\star}(x)\!=\!
\argmax_{\|\Delta\|_2\le \varepsilon_{\mathrm{intra}}}\!
\mathrm{KL}\big(
p_\theta(\cdot\mid x)\,\|\,p_\theta(\cdot\mid x,\Delta)
\big),
\end{equation}
and the outer loss is:
\begin{equation}
\label{eq:intra_trades_outer}
\mathcal{L}_{\mathrm{intra}}^{\mathrm{TR}}
=
\mathbb{E}\!\left[
\mathrm{KL}\big(
p_\theta(\cdot\mid x)\,\|\,p_\theta(\cdot\mid x,\Delta_{\mathrm{intra}}^{\star}(x))
\big)
\right].
\end{equation}

\subsection{Similarity Clipping (Proto-Clip)}
Proto-Clip discourages the model from overly strong alignment to the current task prototype $c_t\in\mathbb{R}^{d}$ via a similarity clipping loss. For the current batch $\mathcal{B}$, mean representation is:
\begin{equation}
\label{eq:batch_mean}
\bar{h}
=
\frac{1}{|\mathcal{B}|}\sum_{x} h_\theta(x),\ x\in\mathcal{B}.
\end{equation}

The prototype is updated by an exponential moving average:
\begin{equation}
\label{eq:proto_ema}
c_t \leftarrow m\,c_t + (1-m)\,\bar{h},
\end{equation}
where $m\in(0,1)$ is the momentum coefficient.

Based on $c_t$, we introduce a penalty that enforces an upper bound on similarity:
\begin{equation}
\label{eq:cut_loss}
\mathcal{L}_{\mathrm{clip}}
=
\mathbb{E}\!\left[
\max\!\big(0,\ S(h_\theta(x),c_t)-\tau_{\mathrm{hi}}\big)
\right]^2,
\end{equation}
where $\tau_{\mathrm{hi}}$ is a similarity threshold. This term acts as a soft constraint on overly high similarity and is activated only when $S(h_\theta(x),c_t)>\tau_{\mathrm{hi}}$.

\subsection{Inter-Task Directional Alignment (Inter-Align)}
Inter-Align enhances retention and stability by shrinking the geometric gap between adjacent tasks. We treat the previous task prototype $c_{t-1}$ as a semantic anchor and load it at the start of task $t$. From the current batch $\mathcal{B}$, we sample a subset $\mathcal{B}_{\mathrm{inter}}$. For any $x\in\mathcal{B}_{\mathrm{inter}}$, the inner perturbation in embedding space is:
\begin{equation}
\label{eq:inter_inner}
\Delta_{\mathrm{inter}}^{\star}(x)\!=\!
\argmax_{\|\Delta\|_2\le \varepsilon_{\mathrm{inter}}}\!
S\big(h_\theta(E(x)+\Delta),c_{t-1}\big).
\end{equation}

This inner maximization seeks the most effective directional alignment perturbation. Maximizing $S(h_\theta(E(x)+\Delta),c_{t-1})$ increases alignment to anchor $c_{t-1}$ and narrows local geometric gap to the previous prototype. It is solved in the same manner as Equation~\ref{eq:intra_pgd}, but with similarity objective in Equation~\ref{eq:inter_inner} replacing $\mathcal{L}_{\mathrm{sup}}$, and using $\varepsilon_{\mathrm{inter}}$ and $K_{\mathrm{inter}}$. After obtaining $\Delta_{\mathrm{inter}}^{\star}(x)$, we compute adversarial loss on perturbed embeddings:
\begin{equation}
\label{eq:inter_outer}
\mathcal{L}_{\mathrm{inter}}
=
\mathbb{E}\!\left[
\mathcal{L}_{\mathrm{sup}}\big(f_\theta(E(x)+\Delta_{\mathrm{inter}}^{\star}(x)),y\big)
\right].
\end{equation}

\subsection{Overall Objective}
For each batch $\mathcal{B}$, we sample $\mathcal{B}_{\mathrm{std}}$, $\mathcal{B}_{\mathrm{intra}}$, and $\mathcal{B}_{\mathrm{inter}}$ with fixed ratios. Let $\mathcal{L}_{\mathrm{std}}$ denote the standard supervised loss over $\mathcal{B}_{\mathrm{std}}$:
\begin{equation}
\label{eq:std_loss}
\mathcal{L}_{\mathrm{std}}
=
\mathbb{E}\!\left[
\mathcal{L}_{\mathrm{sup}}\big(f_\theta(E(x)),y\big)
\right].
\end{equation}

Putting the terms together, the overall objective is:
\begin{equation}
\label{eq:overall_obj}
\begin{aligned}
\mathcal{L}(\theta)
&= \mathcal{L}_{\mathrm{std}}
+ \lambda_{\mathrm{intra}}\mathcal{L}_{\mathrm{intra}} \\
&\quad + \lambda_{\mathrm{clip}}\mathcal{L}_{\mathrm{clip}}
+ \lambda_{\mathrm{inter}}\mathcal{L}_{\mathrm{inter}} .
\end{aligned}
\end{equation}

Except for task prototypes $\{c_t\}$, AdvCL stores neither replay data nor historical models. Adversarial perturbations are generated online during training and are not stored. For robustness evaluation, perturbations are regenerated following evaluation settings.

\section{Experimental Setup}
\subsection{Datasets}
We follow Super Natural Instructions \cite{wang-etal-2022-super} and construct a CL benchmark with six tasks, spanning both classification and generation: tweet anger detection from SemEval 2018 Task 1 \cite{mohammad-etal-2018-semeval}, emotion classification \cite{saravia-etal-2018-carer}, news topic classification on AG News \cite{NIPS2015_250cf8b5}, adversarial natural language inference on ANLI \cite{nie-etal-2020-adversarial}, extreme summarization on XSum \cite{narayan-etal-2018-dont}, and title generation on XL-Sum \cite{hasan-etal-2021-xl}. These tasks exhibit substantial heterogeneity in cross-task similarity, which we characterize in Appendix~\ref{app:dataset}. The default order is anger $\rightarrow$ emotion $\rightarrow$ agnews $\rightarrow$ anli $\rightarrow$ xsum $\rightarrow$ xlsum. For each task, we randomly sample 2,000 samples for training, 200 for validation, and 200 for testing. 

\subsection{Baselines}
All baselines follow the PEFT-CL setting \cite{coleman2025parameterefficientcontinualfinetuningsurvey}. AdvCL comprises three plug-in modules that can be readily integrated into different CL frameworks, so we evaluate on representative CL frameworks that cover replay, regularization, and dynamic architecture methods. Specifically, we include: (1) \textbf{Sequence}, which sequentially fine tunes a single PEFT module across tasks without any CL mechanism. (2) \textbf{Replay}, which sequentially updates the same PEFT module while replaying a fixed proportion of samples from each previous task at every stage. (3) \textbf{EWC} \cite{doi:10.1073/pnas.1611835114}, which adds a quadratic penalty weighted by parameter importance to constrain PEFT parameters that are critical to past tasks. (4) \textbf{MoE}, inspired by prior work \cite{shazeer2017,araujo-etal-2024-learning}, which maintains multiple PEFT experts and uses a router to combine experts for each input during sequential training. (5) \textbf{Joint}, which trains a single PEFT module on the union of all task data. (6) \textbf{Separate}, which trains an independent PEFT module for each task and evaluates by selecting the corresponding module using the given task identity.

\subsection{Metrics}
We report five CL metrics. \textbf{Average Performance (AP)} averages final performance over all tasks after learning the full sequence, using accuracy for classification tasks and ROUGE-L for generation tasks. \textbf{Forgetting (FGT)} quantifies drop on each earlier task from its best achieved value during training to its final value. \textbf{Forward Transfer (FWT)} evaluates how previously learned knowledge helps future tasks. We report two variants: \textbf{FWT-1} uses base model as a zero-shot reference for each upcoming task, while \textbf{FWT-2} uses a single-task trained PEFT model as reference. \textbf{Backward Transfer (BWT)} measures how learning later tasks affects earlier tasks by comparing final performance to performance right after each task is learned. Detailed definitions are provided in Appendix~\ref{app:metrics}.

\subsection{Training Settings}
All experiments run on a single NVIDIA A100 40GB GPU. We use Llama~3.2~3B Instruct as the backbone and adopt LoRA \cite{hu2022lora} as the PEFT module.

For classification tasks, micro batch size is 16 with gradient accumulation of 8, yielding an effective batch size of 128. For generation tasks, micro batch size is 4 with gradient accumulation of 32. Learning rate is set to 3e-5 with a warmup ratio of 0.05 and weight decay 0.1. Baseline-specific settings and hyperparameters of three modules are provided in Appendix~\ref{app:training details}.

\begin{table*}[t]
\centering
\caption{\label{tab:main_standard}
Results of module ablation and transferability under standard setting.
}
\begin{tabular}{lccccc}
\toprule
\textbf{Model} & \textbf{AP($\uparrow$)} & \textbf{FGT($\downarrow$)} & \textbf{FWT-1($\uparrow$)} & \textbf{FWT-2($\uparrow$)} & \textbf{BWT($\uparrow$)} \\
\midrule[0.9pt]
\multicolumn{6}{c}{\textbf{Part I: Module Ablation}} \\
\midrule[0.9pt]
\textbf{Sequence}
& 55.21 & 3.42 & 1.65 & 1.76 & -2.11 \\
+ S (Smooth)
& 55.90 & 2.70 & \uline{2.70} & 1.57 & -1.05 \\
+ C (Clip)
& 55.29 & 3.38 & -0.42 & \uline{1.99} & -2.28 \\
+ A (Align)
& 55.36 & 2.20 & 0.28 & 1.23 & -1.29 \\
+ S + C
& \uline{55.97} & 2.30 & 2.52 & 1.10 & \uline{-0.41} \\
+ S + A
& 55.71 & \uline{1.80} & \textbf{3.55} & 1.06 & -0.66 \\
+ C + A
& 54.98 & 1.95 & 2.11 & 0.48 & -0.85 \\
+ S + C + A
& \textbf{56.64} & \textbf{1.50} & 2.38 & \textbf{2.73} & \textbf{0.38} \\
\midrule[0.9pt]
\multicolumn{6}{c}{\textbf{Part II: Transferability}} \\
\midrule[0.9pt]
\textbf{Replay}
& 55.61 & 1.70 & \uline{0.35} & 0.46 & \uline{-0.07} \\
+ S (Smooth)
& \textbf{57.89} & \textbf{1.30} & \textbf{3.72} & \textbf{2.99} & -0.37 \\
+ C (Clip)
& \uline{56.09} & \uline{1.33} & -0.54 & 0.74 & \textbf{0.17} \\
+ A (Align)
& 55.85 & 2.00 & -1.46 & \uline{1.47} & -1.00 \\
\midrule[0.9pt]
\textbf{EWC}
& 55.42 & 3.51 & -0.06 & \textbf{1.97} & -2.11 \\
+ S (Smooth)
& \uline{55.66} & 3.40 & \textbf{3.05} & \uline{1.76} & -1.57 \\
+ C (Clip)
& 55.61 & \uline{2.56} & \uline{0.83} & 1.28 & \uline{-1.06} \\
+ A (Align)
& \textbf{55.86} & \textbf{1.80} & -0.08 & 1.08 & \textbf{-0.52} \\
\midrule[0.9pt]
\textbf{MoE}
& 53.42 & 4.80 & \textbf{0.46} & 0.93 & -3.27 \\
+ S (Smooth)
& \textbf{54.26} & \textbf{3.30} & -1.31 & 0.86 & \textbf{-2.16} \\
+ C (Clip)
& 53.84 & \uline{4.20} & \uline{0.21} & \uline{1.26} & -3.15 \\
+ A (Align)
& \uline{54.12} & 4.40 & -0.21 & \textbf{1.38} & \uline{-2.97} \\
\midrule[0.9pt]
\textbf{Joint}
& 52.92 & - & - & - & - \\
+ S (Smooth)
& \textbf{56.27} & - & - & - & - \\
\midrule[0.9pt]
\textbf{Separate}
& 55.20 & - & - & - & - \\
+ S (Smooth)
& \textbf{56.76} & - & - & - & - \\
\bottomrule
\end{tabular}
\end{table*}

\section{Main Results}
This section evaluates standard performance across baselines and their module-augmented variants. We present results on module ablation and cross-baseline transferability under the standard setting, as shown in Table~\ref{tab:main_standard}, where S + C + A denotes AdvCL. Within each model family, the best and second-best results are highlighted in bold and underlined, respectively. 

\subsection{Module Ablation}
See Part I of Table~\ref{tab:main_standard} for results. Intra-Smooth trains the model along locally challenging directions in embedding space and serves as a major contributor to overall performance gains. Inter-Align explicitly aligns representations across tasks, which reduces representation drift and improves retention, as reflected by lower forgetting and higher BWT, while its effect on forward transfer is more limited. Proto-Clip alone offers relatively limited gains, but when combined with Intra-Smooth, it strengthens retention-related performance, especially BWT. Intra-Smooth plus Inter-Align is more effective at reducing forgetting and improving FWT-1. Combining all three modules achieves the best results on four metrics, indicating a better trade-off among accuracy, transfer, and forgetting. Additional order results are provided in Appendix~\ref{app:additional order}.

In summary, \textbf{the three-module combination is the most stable, and two-module combinations generally outperform single-module ones}, reflecting clear complementarity among the modules. The main exception is Proto-Clip plus Inter-Align, where both mechanisms impose additional constraints on optimization and may therefore limit the attainable gains.

\subsection{Transferability}
See Part II of Table~\ref{tab:main_standard} for results. To isolate transferability, we add each module individually to each baseline. All three modules exhibit strong transferability, without any performance collapse, and they typically improve performance. In particular, Intra-Smooth is the most consistently beneficial module across baselines.

Proto-Clip and Inter-Align primarily influence stability, but with different patterns. Proto-Clip penalizes over-alignment to the current prototype and is most effective when paired with other modules. Inter-Align more directly shrinks cross-task representational gaps and shows more consistent retention benefits, typically reflected in lower forgetting and higher BWT.

Since all unaugmented baselines share the same backbone and training steps, the aggregate metric AP varies only slightly. In contrast, CL metrics show clearer separation. For example, Replay exhibits substantially lower forgetting. Joint tends to achieve lower AP, which may reflect optimization difficulties caused by multi-task objective conflicts or distributional mismatch. Overall, \textbf{the modules are plug-and-play and do not depend on specific CL details across training paradigms}.

\section{Further Analysis}

\begin{table*}[htbp]
\centering
\caption{Cross-attack results averaged over five attacks.}
\label{tab:cross_attack_avg}
\begin{tabular}{l|ccccc}
\toprule
\textbf{Model} & \textbf{AP$(\uparrow)$} & \textbf{FGT$(\downarrow)$} & \textbf{FWT-1$(\uparrow)$} & \textbf{FWT-2$(\uparrow)$} & \textbf{BWT$(\uparrow)$} \\
\midrule
\textbf{Sequence} & 37.47 & 8.78 & 2.20 & 1.61 & -7.52 \\
+ S (Smooth) & 43.23 & 4.04 & \textbf{3.72} & 2.50 & -1.68 \\
+ C (Clip) & 36.24 & 9.76 & 1.59 & 0.67 & -7.87 \\
+ A (Align) & 39.33 & 5.14 & 1.67 & -0.38 & -3.46 \\
+ S + C + A & \textbf{44.22} & \textbf{2.76} & 3.48 & \textbf{2.90} & \textbf{-0.40} \\
\bottomrule
\end{tabular}
\end{table*}

\subsection{Robustness Evaluation}
We further conduct cross-attack evaluation under five attacks (PGD-5, PGD-10, FGSM, Rand, and discrete perturbations), with averaged results reported in Table~\ref{tab:cross_attack_avg}. Intra-Smooth yields consistent robustness gains, indicating strong generalization. Inter-Align primarily improves stability metrics, especially FGT and BWT. Proto-Clip alone degrades robustness under this evaluation, suggesting an overly restrictive constraint. In contrast, combining all three modules achieves best performance on all metrics, highlighting clear synergistic effects. Additional baselines are provided in Appendix~\ref{app:cross}. Robustness under PGD with varying perturbation strengths is also reported in Appendix~\ref{app:robust pgd} and shows a similar overall pattern.

\begin{table*}[t]
\centering
\caption{Standard evaluation on Sequence baseline with Intra-Smooth. Left: sensitivity to perturbation strength $\varepsilon_{\mathrm{intra}}$. Right: comparison of different solvers used to generate adversarial perturbations. FWT is computed by averaging FWT-1 and FWT-2. Finer-grained $\varepsilon_{\mathrm{intra}}$ values show the same trend and are omitted for brevity. The suffix ``-$K$'' indicates using $K_{\mathrm{intra}}$ steps to generate perturbations, and ``All'' means perturbing both instruction and input embeddings.}

\label{tab:intra-sensitivity}
\begin{tabular}{lcccc@{\hspace{4pt}}|@{\hspace{4pt}}lcccc}
\toprule
\multicolumn{5}{c}{\textbf{Part I: Perturbation Strength ($\varepsilon_{\mathrm{intra}}$)}} &
\multicolumn{5}{c}{\textbf{Part II: Perturbation Solver}} \\
\cmidrule(r){1-5}\cmidrule(l){6-10}
\textbf{Setting}
& \shortstack{\textbf{AP}\\($\uparrow$)}
& \shortstack{\textbf{FGT}\\($\downarrow$)}
& \shortstack{\textbf{FWT}\\($\uparrow$)}
& \shortstack{\textbf{BWT}\\($\uparrow$)}
& \textbf{Setting}
& \shortstack{\textbf{AP}\\($\uparrow$)}
& \shortstack{\textbf{FGT}\\($\downarrow$)}
& \shortstack{\textbf{FWT}\\($\uparrow$)}
& \shortstack{\textbf{BWT}\\($\uparrow$)} \\
\midrule
0.00 & 55.21 & 3.42 & 1.71 & -2.11
& PGD-1 & 55.90 & 2.70 & 2.14 & -1.05 \\
0.05 & 55.75 & 2.70 & 2.16 & -1.59
& PGD-3 & 55.81 & 3.56 & 3.17 & -3.16 \\
0.10 & 55.90 & 2.70 & 2.14 & -1.05
& PGD-10 & 55.94 & 3.93 & 3.26 & -3.53 \\
0.15 & 53.87 & 6.04 & 2.55 & -4.94
& TRADES-1 & 55.78 & 3.17 & 1.41 & -1.97 \\
0.20 & 54.78 & 4.90 & 0.90 & -3.25
& FGSM & 56.30 & 3.03 & 2.00 & -1.53 \\
0.25 & 55.70 & 3.06 & 0.49 & -1.56
& Rand & 55.14 & 3.38 & 0.58 & -1.78 \\
0.30 & 56.66 & 1.55 & 0.22 & -0.05
& PGD-1 All & 55.05 & 4.46 & 2.29 & -3.16 \\
\bottomrule
\end{tabular}
\end{table*}

\subsection{Intra-Smooth Sensitivity}
\subsubsection{Perturbation Strength.} Part I of Table \ref{tab:intra-sensitivity} shows the effect of different perturbation strengths in Intra-Smooth. FGT and BWT vary non-monotonically with $\varepsilon_{\mathrm{intra}}$. Small perturbations ($0.05$, $0.10$) behave like a mild regularizer for local smoothing, reducing forgetting and improving BWT. Intermediate strengths ($0.15$, $0.20$) form a transition region, where perturbations amplify conflicts with stability on clean samples, but do not yield a sufficiently consistent smoothing signal. As a result, stability degrades and AP drops. With larger perturbations ($0.25$, $0.30$), adversarial examples become consistently harder across samples, which strengthens the smoothing effect and restores stability, with AP recovering accordingly.

In contrast, FWT decreases as $\varepsilon_{\mathrm{intra}}$ grows, suggesting that stronger smoothing constrains effective updates and favors neighborhood consistency over forward transfer.

\subsubsection{Perturbation Solver.} Part II of Table~\ref{tab:intra-sensitivity} compares different perturbation solvers. Within PGD-$K$ family, increasing $K$ reveals a clear trade-off between stability and plasticity. Larger $K$ tends to raise FWT, while retention metrics can degrade, with higher forgetting and a more negative BWT. Compared with PGD-1, alternative solvers do not offer a consistent improvement across metrics. PGD-1 All perturbs both instruction and input tokens, which often hurts stability. Rand lacks worst-case optimization and therefore offers little benefit. TRADES-1 and FGSM achieve AP comparable to PGD-1, but their retention metrics are less reliable than it. TRADES-1 optimizes a different objective, while FGSM follows a different perturbation constraint. Empirically, PGD-1 is a reliable and well-balanced default choice across metrics.

\subsection{Inter-Align Geometry}

To examine how Inter-Align affects representation geometry across tasks, we quantify geometric relationship between adjacent tasks in representation space using three complementary metrics: (1) \textbf{Centroid Cosine Distance} measures global gap between representation centroids of previous and current tasks; (2) \textbf{Mean Pairwise Cosine Distance} measures distributional gap by averaging cosine distances over cross-task sample pairs; and (3) \textbf{Directional Prototype Alignment} measures alignment strength between current task representations and previous task prototype direction. Using representations extracted after training each task, Inter-Align yields average changes of -4.35\%, -0.61\%, and +0.33\% on these three metrics, respectively, compared with sequence training over adjacent task pairs. These results suggest that Inter-Align narrows cross-task geometric gaps and strengthens directional alignment. See Appendix~\ref{app:geometric metrics} for detailed metric definitions and results.

Further analysis of similarity shift under directional perturbations, together with the impact of $\varepsilon_{\mathrm{inter}}$ on continual learning metrics, is provided in Appendix~\ref{app:inter sensitivity}.

\subsection{Computational Overhead Analysis}
To quantify computational cost of each module, we compare their additional training overhead under default settings. Intra-Smooth and Inter-Align incur approximately +25\% and +12.5\% additional computation, respectively. Proto-Clip is computed using activations from baseline training step and thus introduces negligible extra computation. Overall, the computational overhead remains predictable and manageable. Three-module combination adds about +37.5\% computation, which is justified given the gains. More details are provided in Appendix~\ref{app:overhead}.

\subsection{Additional Performance Analysis}
We further analyze the proposed method from two perspectives: per-task stagewise behavior and the relationship between standard and robust performance. Detailed results are provided in Appendix~\ref{app:stagewise and tradeoff}. First, the per-task stagewise performance curves show that the three-module combination mainly supports long-term stability and transfer across stages, rather than maximizing performance at each stage. Although the gains differ across tasks, it generally leads to more stable trajectories and better preservation of earlier-task performance in later stages.

Second, we compare standard and robust average performance under two PGD-based evaluation settings. Results suggest that standard and robust performance do not follow a simple trade-off. In most cases, adding our modules improves robustness while maintaining competitive or better standard performance. Intra-Smooth provides the most consistent gains, while the full combination remains strong overall.

\section{Conclusion}
We propose AdvCL, which repurposes adversarial perturbations as a geometric control signal for CL. AdvCL includes three transferable plug-in modules, Intra-Smooth promotes local smoothing and robustness, Proto-Clip prevents over-alignment to current prototype, and Inter-Align reduces cross-task representational gaps via directional alignment. Experiments show consistent gains in both standard performance and robustness, with reduced forgetting and better transfer.


\bibliography{custom}

@inproceedings{lester-etal-2021-power,
    title = "The Power of Scale for Parameter-Efficient Prompt Tuning",
    author = "Lester, Brian  and
      Al-Rfou, Rami  and
      Constant, Noah",
    editor = "Moens, Marie-Francine  and
      Huang, Xuanjing  and
      Specia, Lucia  and
      Yih, Scott Wen-tau",
    booktitle = "Proceedings of the 2021 Conference on Empirical Methods in Natural Language Processing",
    month = nov,
    year = "2021",
    address = "Online and Punta Cana, Dominican Republic",
    publisher = "Association for Computational Linguistics",
    doi = "10.18653/v1/2021.emnlp-main.243",
    pages = "3045--3059"
}

@inproceedings{ji-etal-2025-language-models,
    title = "Language Models Resist Alignment: Evidence From Data Compression",
    author = "Ji, Jiaming  and
      Wang, Kaile  and
      Qiu, Tianyi Alex  and
      Chen, Boyuan  and
      Zhou, Jiayi  and
      Li, Changye  and
      Lou, Hantao  and
      Dai, Josef  and
      Liu, Yunhuai  and
      Yang, Yaodong",
    editor = "Che, Wanxiang  and
      Nabende, Joyce  and
      Shutova, Ekaterina  and
      Pilehvar, Mohammad Taher",
    booktitle = "Proceedings of the 63rd Annual Meeting of the Association for Computational Linguistics (Volume 1: Long Papers)",
    month = jul,
    year = "2025",
    address = "Vienna, Austria",
    publisher = "Association for Computational Linguistics",
    doi = "10.18653/v1/2025.acl-long.1141",
    pages = "23411--23432",
    ISBN = "979-8-89176-251-0"
}

@inproceedings{
keskar2017on,
title={On Large-Batch Training for Deep Learning: Generalization Gap and Sharp Minima},
author={Nitish Shirish Keskar and Dheevatsa Mudigere and Jorge Nocedal and Mikhail Smelyanskiy and Ping Tak Peter Tang},
booktitle={International Conference on Learning Representations},
year={2017},
}

@InProceedings{pmlr-v97-zhang19p,
  title = 	 {Theoretically Principled Trade-off between Robustness and Accuracy},
  author =       {Zhang, Hongyang and Yu, Yaodong and Jiao, Jiantao and Xing, Eric and Ghaoui, Laurent El and Jordan, Michael},
  booktitle = 	 {Proceedings of the 36th International Conference on Machine Learning},
  pages = 	 {7472--7482},
  year = 	 {2019},
  editor = 	 {Chaudhuri, Kamalika and Salakhutdinov, Ruslan},
  volume = 	 {97},
  series = 	 {Proceedings of Machine Learning Research},
  month = 	 {09--15 Jun},
  publisher =    {PMLR}
}

@misc{ru2024maintainingadversarialrobustnesscontinuous,
      title={Maintaining Adversarial Robustness in Continuous Learning}, 
      author={Xiaolei Ru and Xiaowei Cao and Zijia Liu and Jack Murdoch Moore and Xin-Ya Zhang and Xia Zhu and Wenjia Wei and Gang Yan},
      year={2024},
      eprint={2402.11196},
      archivePrefix={arXiv},
      primaryClass={cs.LG},
}

@inproceedings{NEURIPS2024_05cdc7fe,
 author = {Hiratani, Naoki},
 booktitle = {Advances in Neural Information Processing Systems},
 doi = {10.52202/079017-0107},
 editor = {A. Globerson and L. Mackey and D. Belgrave and A. Fan and U. Paquet and J. Tomczak and C. Zhang},
 pages = {3243--3274},
 publisher = {Curran Associates, Inc.},
 title = {Disentangling and mitigating the impact of task similarity for continual learning},
 volume = {37},
 year = {2024}
}

@InProceedings{Achille_2019_ICCV,
author = {Achille, Alessandro and Lam, Michael and Tewari, Rahul and Ravichandran, Avinash and Maji, Subhransu and Fowlkes, Charless C. and Soatto, Stefano and Perona, Pietro},
title = {Task2Vec: Task Embedding for Meta-Learning},
booktitle = {Proceedings of the IEEE/CVF International Conference on Computer Vision (ICCV)},
month = {October},
year = {2019}
}

@InProceedings{Rebuffi_2017_CVPR,
author = {Rebuffi, Sylvestre-Alvise and Kolesnikov, Alexander and Sperl, Georg and Lampert, Christoph H.},
title = {iCaRL: Incremental Classifier and Representation Learning},
booktitle = {Proceedings of the IEEE Conference on Computer Vision and Pattern Recognition (CVPR)},
month = {July},
year = {2017}
}

@InProceedings{Cha_2021_ICCV,
    author    = {Cha, Hyuntak and Lee, Jaeho and Shin, Jinwoo},
    title     = {Co2L: Contrastive Continual Learning},
    booktitle = {Proceedings of the IEEE/CVF International Conference on Computer Vision (ICCV)},
    month     = {October},
    year      = {2021},
    pages     = {9516-9525}
}

@ARTICLE{8417973,
  author={Miyato, Takeru and Maeda, Shin-Ichi and Koyama, Masanori and Ishii, Shin},
  journal={IEEE Transactions on Pattern Analysis and Machine Intelligence}, 
  title={Virtual Adversarial Training: A Regularization Method for Supervised and Semi-Supervised Learning}, 
  year={2019},
  volume={41},
  number={8},
  pages={1979-1993},
  doi={10.1109/TPAMI.2018.2858821}}

@inproceedings{
foret2021sharpnessaware,
title={Sharpness-aware Minimization for Efficiently Improving Generalization},
author={Pierre Foret and Ariel Kleiner and Hossein Mobahi and Behnam Neyshabur},
booktitle={International Conference on Learning Representations},
year={2021},
}

@inproceedings{huang-etal-2024-mitigating,
    title = "Mitigating Catastrophic Forgetting in Large Language Models with Self-Synthesized Rehearsal",
    author = "Huang, Jianheng  and
      Cui, Leyang  and
      Wang, Ante  and
      Yang, Chengyi  and
      Liao, Xinting  and
      Song, Linfeng  and
      Yao, Junfeng  and
      Su, Jinsong",
    editor = "Ku, Lun-Wei  and
      Martins, Andre  and
      Srikumar, Vivek",
    booktitle = "Proceedings of the 62nd Annual Meeting of the Association for Computational Linguistics (Volume 1: Long Papers)",
    month = aug,
    year = "2024",
    address = "Bangkok, Thailand",
    publisher = "Association for Computational Linguistics",
    doi = "10.18653/v1/2024.acl-long.77",
    pages = "1416--1428"
}

@misc{rusu2022progressiveneuralnetworks,
      title={Progressive Neural Networks}, 
      author={Andrei A. Rusu and Neil C. Rabinowitz and Guillaume Desjardins and Hubert Soyer and James Kirkpatrick and Koray Kavukcuoglu and Razvan Pascanu and Raia Hadsell},
      year={2016},
      eprint={1606.04671},
      archivePrefix={arXiv},
      primaryClass={cs.LG},
}

@inproceedings{wang-etal-2024-rehearsal,
    title = "Rehearsal-Free Modular and Compositional Continual Learning for Language Models",
    author = {Wang, Mingyang  and
      Adel, Heike  and
      Lange, Lukas  and
      Str{\"o}tgen, Jannik  and
      Schuetze, Hinrich},
    editor = "Duh, Kevin  and
      Gomez, Helena  and
      Bethard, Steven",
    booktitle = "Proceedings of the 2024 Conference of the North American Chapter of the Association for Computational Linguistics: Human Language Technologies (Volume 2: Short Papers)",
    month = jun,
    year = "2024",
    address = "Mexico City, Mexico",
    publisher = "Association for Computational Linguistics",
    doi = "10.18653/v1/2024.naacl-short.39",
    pages = "469--480"
}

@article{
doi:10.1073/pnas.1611835114,
author = {James Kirkpatrick  and Razvan Pascanu  and Neil Rabinowitz  and Joel Veness  and Guillaume Desjardins  and Andrei A. Rusu  and Kieran Milan  and John Quan  and Tiago Ramalho  and Agnieszka Grabska-Barwinska  and Demis Hassabis  and Claudia Clopath  and Dharshan Kumaran  and Raia Hadsell },
title = {Overcoming catastrophic forgetting in neural networks},
journal = {Proceedings of the National Academy of Sciences},
volume = {114},
number = {13},
pages = {3521-3526},
year = {2017},
doi = {10.1073/pnas.1611835114},
eprint = {https://www.pnas.org/doi/pdf/10.1073/pnas.1611835114}}

@inproceedings{wang-etal-2023-orthogonal,
    title = "Orthogonal Subspace Learning for Language Model Continual Learning",
    author = "Wang, Xiao  and
      Chen, Tianze  and
      Ge, Qiming  and
      Xia, Han  and
      Bao, Rong  and
      Zheng, Rui  and
      Zhang, Qi  and
      Gui, Tao  and
      Huang, Xuanjing",
    editor = "Bouamor, Houda  and
      Pino, Juan  and
      Bali, Kalika",
    booktitle = "Findings of the Association for Computational Linguistics: EMNLP 2023",
    month = dec,
    year = "2023",
    address = "Singapore",
    publisher = "Association for Computational Linguistics",
    doi = "10.18653/v1/2023.findings-emnlp.715",
    pages = "10658--10671"
}

@inproceedings{lu-etal-2025-controlled,
    title = "Controlled Low-Rank Adaptation with Subspace Regularization for Continued Training on Large Language Models",
    author = "Lu, Yuheng  and
      Qian, Bingshuo  and
      Yuan, Caixia  and
      Jiang, Huixing  and
      Wang, Xiaojie",
    editor = "Che, Wanxiang  and
      Nabende, Joyce  and
      Shutova, Ekaterina  and
      Pilehvar, Mohammad Taher",
    booktitle = "Proceedings of the 63rd Annual Meeting of the Association for Computational Linguistics (Volume 1: Long Papers)",
    month = jul,
    year = "2025",
    address = "Vienna, Austria",
    publisher = "Association for Computational Linguistics",
    doi = "10.18653/v1/2025.acl-long.940",
    pages = "19165--19181",
    ISBN = "979-8-89176-251-0"
}

@inproceedings{
hu2022lora,
title={Lo{RA}: Low-Rank Adaptation of Large Language Models},
author={Edward J Hu and Yelong Shen and Phillip Wallis and Zeyuan Allen-Zhu and Yuanzhi Li and Shean Wang and Lu Wang and Weizhu Chen},
booktitle={International Conference on Learning Representations},
year={2022},
}

@inproceedings{zhao-etal-2024-sapt,
    title = "{SAPT}: A Shared Attention Framework for Parameter-Efficient Continual Learning of Large Language Models",
    author = "Zhao, Weixiang  and
      Wang, Shilong  and
      Hu, Yulin  and
      Zhao, Yanyan  and
      Qin, Bing  and
      Zhang, Xuanyu  and
      Yang, Qing  and
      Xu, Dongliang  and
      Che, Wanxiang",
    editor = "Ku, Lun-Wei  and
      Martins, Andre  and
      Srikumar, Vivek",
    booktitle = "Proceedings of the 62nd Annual Meeting of the Association for Computational Linguistics (Volume 1: Long Papers)",
    month = aug,
    year = "2024",
    address = "Bangkok, Thailand",
    publisher = "Association for Computational Linguistics",
    doi = "10.18653/v1/2024.acl-long.625",
    pages = "11641--11661"
}

@inproceedings{han-etal-2025-slim,
    title = "{SLIM}: Let {LLM} Learn More and Forget Less with Soft {L}o{RA} and Identity Mixture",
    author = "Han, Jiayi  and
      Du, Liang  and
      Du, Hongwei  and
      Zhou, Xiangguo  and
      Wu, Yiwen  and
      Zhang, Yuanfang  and
      Zheng, Weibo  and
      Han, Donghong",
    editor = "Chiruzzo, Luis  and
      Ritter, Alan  and
      Wang, Lu",
    booktitle = "Proceedings of the 2025 Conference of the Nations of the Americas Chapter of the Association for Computational Linguistics: Human Language Technologies (Volume 1: Long Papers)",
    month = apr,
    year = "2025",
    address = "Albuquerque, New Mexico",
    publisher = "Association for Computational Linguistics",
    doi = "10.18653/v1/2025.naacl-long.246",
    pages = "4792--4804",
    ISBN = "979-8-89176-189-6"
}

@InProceedings{pmlr-v97-houlsby19a,
  title = 	 {Parameter-Efficient Transfer Learning for {NLP}},
  author =       {Houlsby, Neil and Giurgiu, Andrei and Jastrzebski, Stanislaw and Morrone, Bruna and De Laroussilhe, Quentin and Gesmundo, Andrea and Attariyan, Mona and Gelly, Sylvain},
  booktitle = 	 {Proceedings of the 36th International Conference on Machine Learning},
  pages = 	 {2790--2799},
  year = 	 {2019},
  editor = 	 {Chaudhuri, Kamalika and Salakhutdinov, Ruslan},
  volume = 	 {97},
  series = 	 {Proceedings of Machine Learning Research},
  month = 	 {09--15 Jun},
  publisher =    {PMLR}
}

@inproceedings{
Zhu2020FreeLB,
title={FreeLB: Enhanced Adversarial Training for Natural Language Understanding},
author={Chen Zhu and Yu Cheng and Zhe Gan and Siqi Sun and Tom Goldstein and Jingjing Liu},
booktitle={International Conference on Learning Representations},
year={2020},
}

@inproceedings{jiang-etal-2020-smart,
    title = "{SMART}: Robust and Efficient Fine-Tuning for Pre-trained Natural Language Models through Principled Regularized Optimization",
    author = "Jiang, Haoming  and
      He, Pengcheng  and
      Chen, Weizhu  and
      Liu, Xiaodong  and
      Gao, Jianfeng  and
      Zhao, Tuo",
    editor = "Jurafsky, Dan  and
      Chai, Joyce  and
      Schluter, Natalie  and
      Tetreault, Joel",
    booktitle = "Proceedings of the 58th Annual Meeting of the Association for Computational Linguistics",
    month = jul,
    year = "2020",
    address = "Online",
    publisher = "Association for Computational Linguistics",
    doi = "10.18653/v1/2020.acl-main.197",
    pages = "2177--2190"
}

@inproceedings{
aghajanyan2021better,
title={Better Fine-Tuning by Reducing Representational Collapse},
author={Armen Aghajanyan and Akshat Shrivastava and Anchit Gupta and Naman Goyal and Luke Zettlemoyer and Sonal Gupta},
booktitle={International Conference on Learning Representations},
year={2021},
}

@inproceedings{
jain2024neftune,
title={{NEFT}une: Noisy Embeddings Improve Instruction Finetuning},
author={Neel Jain and Ping{-}yeh Chiang and Yuxin Wen and John Kirchenbauer and Hong-Min Chu and Gowthami Somepalli and Brian R. Bartoldson and Bhavya Kailkhura and Avi Schwarzschild and Aniruddha Saha and Micah Goldblum and Jonas Geiping and Tom Goldstein},
booktitle={The Twelfth International Conference on Learning Representations},
year={2024},
}

@misc{bell2022effecttaskorderingcontinual,
      title={The Effect of Task Ordering in Continual Learning}, 
      author={Samuel J. Bell and Neil D. Lawrence},
      year={2022},
      eprint={2205.13323},
      archivePrefix={arXiv},
      primaryClass={cs.LG},
}

@inproceedings{wang-etal-2019-sentence,
    title = "Sentence Embedding Alignment for Lifelong Relation Extraction",
    author = "Wang, Hong  and
      Xiong, Wenhan  and
      Yu, Mo  and
      Guo, Xiaoxiao  and
      Chang, Shiyu  and
      Wang, William Yang",
    editor = "Burstein, Jill  and
      Doran, Christy  and
      Solorio, Thamar",
    booktitle = "Proceedings of the 2019 Conference of the North {A}merican Chapter of the Association for Computational Linguistics: Human Language Technologies, Volume 1 (Long and Short Papers)",
    month = jun,
    year = "2019",
    address = "Minneapolis, Minnesota",
    publisher = "Association for Computational Linguistics",
    doi = "10.18653/v1/N19-1086",
    pages = "796--806"
}

@inproceedings{cui-etal-2021-refining,
    title = "Refining Sample Embeddings with Relation Prototypes to Enhance Continual Relation Extraction",
    author = "Cui, Li  and
      Yang, Deqing  and
      Yu, Jiaxin  and
      Hu, Chengwei  and
      Cheng, Jiayang  and
      Yi, Jingjie  and
      Xiao, Yanghua",
    editor = "Zong, Chengqing  and
      Xia, Fei  and
      Li, Wenjie  and
      Navigli, Roberto",
    booktitle = "Proceedings of the 59th Annual Meeting of the Association for Computational Linguistics and the 11th International Joint Conference on Natural Language Processing (Volume 1: Long Papers)",
    month = aug,
    year = "2021",
    address = "Online",
    publisher = "Association for Computational Linguistics",
    doi = "10.18653/v1/2021.acl-long.20",
    pages = "232--243"
}

@inproceedings{zhao-etal-2022-consistent,
    title = "Consistent Representation Learning for Continual Relation Extraction",
    author = "Zhao, Kang  and
      Xu, Hua  and
      Yang, Jiangong  and
      Gao, Kai",
    editor = "Muresan, Smaranda  and
      Nakov, Preslav  and
      Villavicencio, Aline",
    booktitle = "Findings of the Association for Computational Linguistics: ACL 2022",
    month = may,
    year = "2022",
    address = "Dublin, Ireland",
    publisher = "Association for Computational Linguistics",
    doi = "10.18653/v1/2022.findings-acl.268",
    pages = "3402--3411"
}

@inproceedings{
madry2018towards,
title={Towards Deep Learning Models Resistant to Adversarial Attacks},
author={Aleksander Madry and Aleksandar Makelov and Ludwig Schmidt and Dimitris Tsipras and Adrian Vladu},
booktitle={International Conference on Learning Representations},
year={2018},
}

@inproceedings{
goodfellow2015explainingharnessingadversarialexamples,
title={Explaining and Harnessing Adversarial Examples},
author={Ian J. Goodfellow and Jonathon Shlens and Christian Szegedy},
booktitle={International Conference on Learning Representations},
year={2015},
}

@inproceedings{wang-etal-2022-super,
    title = "Super-{N}atural{I}nstructions: Generalization via Declarative Instructions on 1600+ {NLP} Tasks",
    author = "Wang, Yizhong  and
      Mishra, Swaroop  and
      Alipoormolabashi, Pegah  and
      Kordi, Yeganeh  and
      Mirzaei, Amirreza  and
      Naik, Atharva  and
      Ashok, Arjun  and
      Dhanasekaran, Arut Selvan  and
      Arunkumar, Anjana  and
      Stap, David  and
      Pathak, Eshaan  and
      Karamanolakis, Giannis  and
      Lai, Haizhi  and
      Purohit, Ishan  and
      Mondal, Ishani  and
      Anderson, Jacob  and
      Kuznia, Kirby  and
      Doshi, Krima  and
      Pal, Kuntal Kumar  and
      Patel, Maitreya  and
      Moradshahi, Mehrad  and
      Parmar, Mihir  and
      Purohit, Mirali  and
      Varshney, Neeraj  and
      Kaza, Phani Rohitha  and
      Verma, Pulkit  and
      Puri, Ravsehaj Singh  and
      Karia, Rushang  and
      Doshi, Savan  and
      Sampat, Shailaja Keyur  and
      Mishra, Siddhartha  and
      Reddy A, Sujan  and
      Patro, Sumanta  and
      Dixit, Tanay  and
      Shen, Xudong",
    editor = "Goldberg, Yoav  and
      Kozareva, Zornitsa  and
      Zhang, Yue",
    booktitle = "Proceedings of the 2022 Conference on Empirical Methods in Natural Language Processing",
    month = dec,
    year = "2022",
    address = "Abu Dhabi, United Arab Emirates",
    publisher = "Association for Computational Linguistics",
    doi = "10.18653/v1/2022.emnlp-main.340",
    pages = "5085--5109"
}

@inproceedings{mohammad-etal-2018-semeval,
    title = "{S}em{E}val-2018 Task 1: Affect in Tweets",
    author = "Mohammad, Saif  and
      Bravo-Marquez, Felipe  and
      Salameh, Mohammad  and
      Kiritchenko, Svetlana",
    editor = "Apidianaki, Marianna  and
      Mohammad, Saif M.  and
      May, Jonathan  and
      Shutova, Ekaterina  and
      Bethard, Steven  and
      Carpuat, Marine",
    booktitle = "Proceedings of the 12th International Workshop on Semantic Evaluation",
    month = jun,
    year = "2018",
    address = "New Orleans, Louisiana",
    publisher = "Association for Computational Linguistics",
    doi = "10.18653/v1/S18-1001",
    pages = "1--17"
}

@inproceedings{saravia-etal-2018-carer,
    title = "{CARER}: Contextualized Affect Representations for Emotion Recognition",
    author = "Saravia, Elvis  and
      Liu, Hsien-Chi Toby  and
      Huang, Yen-Hao  and
      Wu, Junlin  and
      Chen, Yi-Shin",
    editor = "Riloff, Ellen  and
      Chiang, David  and
      Hockenmaier, Julia  and
      Tsujii, Jun{'}ichi",
    booktitle = "Proceedings of the 2018 Conference on Empirical Methods in Natural Language Processing",
    month = oct # "-" # nov,
    year = "2018",
    address = "Brussels, Belgium",
    publisher = "Association for Computational Linguistics",
    doi = "10.18653/v1/D18-1404",
    pages = "3687--3697"
}

@inproceedings{NIPS2015_250cf8b5,
 author = {Zhang, Xiang and Zhao, Junbo and LeCun, Yann},
 booktitle = {Advances in Neural Information Processing Systems},
 editor = {C. Cortes and N. Lawrence and D. Lee and M. Sugiyama and R. Garnett},
 pages = {},
 publisher = {Curran Associates, Inc.},
 title = {Character-level Convolutional Networks for Text Classification},
 volume = {28},
 year = {2015}
}

@inproceedings{nie-etal-2020-adversarial,
    title = "Adversarial {NLI}: A New Benchmark for Natural Language Understanding",
    author = "Nie, Yixin  and
      Williams, Adina  and
      Dinan, Emily  and
      Bansal, Mohit  and
      Weston, Jason  and
      Kiela, Douwe",
    editor = "Jurafsky, Dan  and
      Chai, Joyce  and
      Schluter, Natalie  and
      Tetreault, Joel",
    booktitle = "Proceedings of the 58th Annual Meeting of the Association for Computational Linguistics",
    month = jul,
    year = "2020",
    address = "Online",
    publisher = "Association for Computational Linguistics",
    doi = "10.18653/v1/2020.acl-main.441",
    pages = "4885--4901"
}

@inproceedings{narayan-etal-2018-dont,
    title = "Don{'}t Give Me the Details, Just the Summary! Topic-Aware Convolutional Neural Networks for Extreme Summarization",
    author = "Narayan, Shashi  and
      Cohen, Shay B.  and
      Lapata, Mirella",
    editor = "Riloff, Ellen  and
      Chiang, David  and
      Hockenmaier, Julia  and
      Tsujii, Jun{'}ichi",
    booktitle = "Proceedings of the 2018 Conference on Empirical Methods in Natural Language Processing",
    month = oct # "-" # nov,
    year = "2018",
    address = "Brussels, Belgium",
    publisher = "Association for Computational Linguistics",
    doi = "10.18653/v1/D18-1206",
    pages = "1797--1807"
}

@inproceedings{hasan-etal-2021-xl,
    title = "{XL}-Sum: Large-Scale Multilingual Abstractive Summarization for 44 Languages",
    author = "Hasan, Tahmid  and
      Bhattacharjee, Abhik  and
      Islam, Md. Saiful  and
      Mubasshir, Kazi  and
      Li, Yuan-Fang  and
      Kang, Yong-Bin  and
      Rahman, M. Sohel  and
      Shahriyar, Rifat",
    editor = "Zong, Chengqing  and
      Xia, Fei  and
      Li, Wenjie  and
      Navigli, Roberto",
    booktitle = "Findings of the Association for Computational Linguistics: ACL-IJCNLP 2021",
    month = aug,
    year = "2021",
    address = "Online",
    publisher = "Association for Computational Linguistics",
    doi = "10.18653/v1/2021.findings-acl.413",
    pages = "4693--4703"
}

@inproceedings{
shazeer2017,
title={ Outrageously Large Neural Networks: The Sparsely-Gated Mixture-of-Experts Layer},
author={Noam Shazeer and Azalia Mirhoseini and Krzysztof Maziarz and Andy Davis and Quoc Le and Geoffrey Hinton and Jeff Dean},
booktitle={International Conference on Learning Representations},
year={2017},
}

@misc{coleman2025parameterefficientcontinualfinetuningsurvey,
      title={Parameter-Efficient Continual Fine-Tuning: A Survey}, 
      author={Eric Nuertey Coleman and Luigi Quarantiello and Ziyue Liu and Qinwen Yang and Samrat Mukherjee and Julio Hurtado and Vincenzo Lomonaco},
      year={2025},
      eprint={2504.13822},
      archivePrefix={arXiv},
      primaryClass={cs.LG},
}

@inproceedings{araujo-etal-2024-learning,
    title = "Learning to Route for Dynamic Adapter Composition in Continual Learning with Language Models",
    author = "Araujo, Vladimir  and
      Moens, Marie-Francine  and
      Tuytelaars, Tinne",
    editor = "Al-Onaizan, Yaser  and
      Bansal, Mohit  and
      Chen, Yun-Nung",
    booktitle = "Findings of the Association for Computational Linguistics: EMNLP 2024",
    month = nov,
    year = "2024",
    address = "Miami, Florida, USA",
    publisher = "Association for Computational Linguistics",
    doi = "10.18653/v1/2024.findings-emnlp.38",
    pages = "687--696"
}

@inproceedings{NEURIPS2024_ef626147,
 author = {Wang, Zhenyi and Huang, Heng},
 booktitle = {Advances in Neural Information Processing Systems},
 doi = {10.52202/079017-4215},
 editor = {A. Globerson and L. Mackey and D. Belgrave and A. Fan and U. Paquet and J. Tomczak and C. Zhang},
 pages = {132583--132613},
 publisher = {Curran Associates, Inc.},
 title = {Model Sensitivity Aware Continual Learning},
 volume = {37},
 year = {2024}
}
\newpage
\appendix

\section{Experimental Details}
\subsection{Dataset Details}
\label{app:dataset}

\paragraph{\textbf{Tweet Anger Detection (anger)}.}
A binary classification task for affect detection on social media. Given a tweet, the model predicts whether the tweet expresses anger. The dataset comes from SemEval 2018 Task 1 \cite{mohammad-etal-2018-semeval}.
\textit{Instruction: You are given a tweet. Decide if it expresses anger. Use the following numeric labels: 0 = Not angry; 1 = Angry.}

\paragraph{\textbf{Emotion Classification (emotion)}.}
A multi-class classification task for recognizing the emotion expressed in a sentence. Given a sentence, the model predicts one emotion label. The dataset comes from \cite{saravia-etal-2018-carer}.
\textit{Instruction: You are given a tweet. Identify the emotion it expresses. Use the following numeric labels: 0 = anger; 1 = fear; 2 = joy; 3 = love; 4 = sadness; 5 = surprise.}

\paragraph{\textbf{News Topic Classification (agnews)}.}
A news topic classification task. Given a short news article, the model predicts its topic. The dataset comes from AG News \cite{NIPS2015_250cf8b5}.
\textit{Instruction: You are given a news article. Classify its topic. Use the following numeric labels: 0 = World; 1 = Sports; 2 = Business; 3 = Science/Technology.}

\paragraph{\textbf{Adversarial Natural Language Inference (anli)}.}
A natural language inference task. Each example consists of a premise and a hypothesis, and the model predicts whether the hypothesis entails, contradicts, or is neutral with respect to the premise. The dataset comes from ANLI \cite{nie-etal-2020-adversarial}.
\textit{Instruction: You are given a premise and a hypothesis. Determine their logical relation. Use the following numeric labels: 0 = Contradiction; 1 = Neutral; 2 = Entailment.}

\paragraph{\textbf{Extreme Summarization (xsum)}.}
A conditional generation task. Given a news article, the model generates a single-sentence summary that captures the core point of the article. The dataset comes from XSum \cite{narayan-etal-2018-dont}.
\textit{Instruction: You are given a news article. Write a one-sentence summary.}

\paragraph{\textbf{Title Generation (xlsum)}.}
A conditional generation task. Given a news text, the model generates a concise title. The dataset comes from XL-Sum \cite{hasan-etal-2021-xl}.
\textit{Instruction: You are given a news article. Generate a concise headline.}

\begin{figure*}[t]
  \centering
  \begin{subfigure}[t]{0.49\linewidth}
    \centering
    \includegraphics[width=\linewidth]{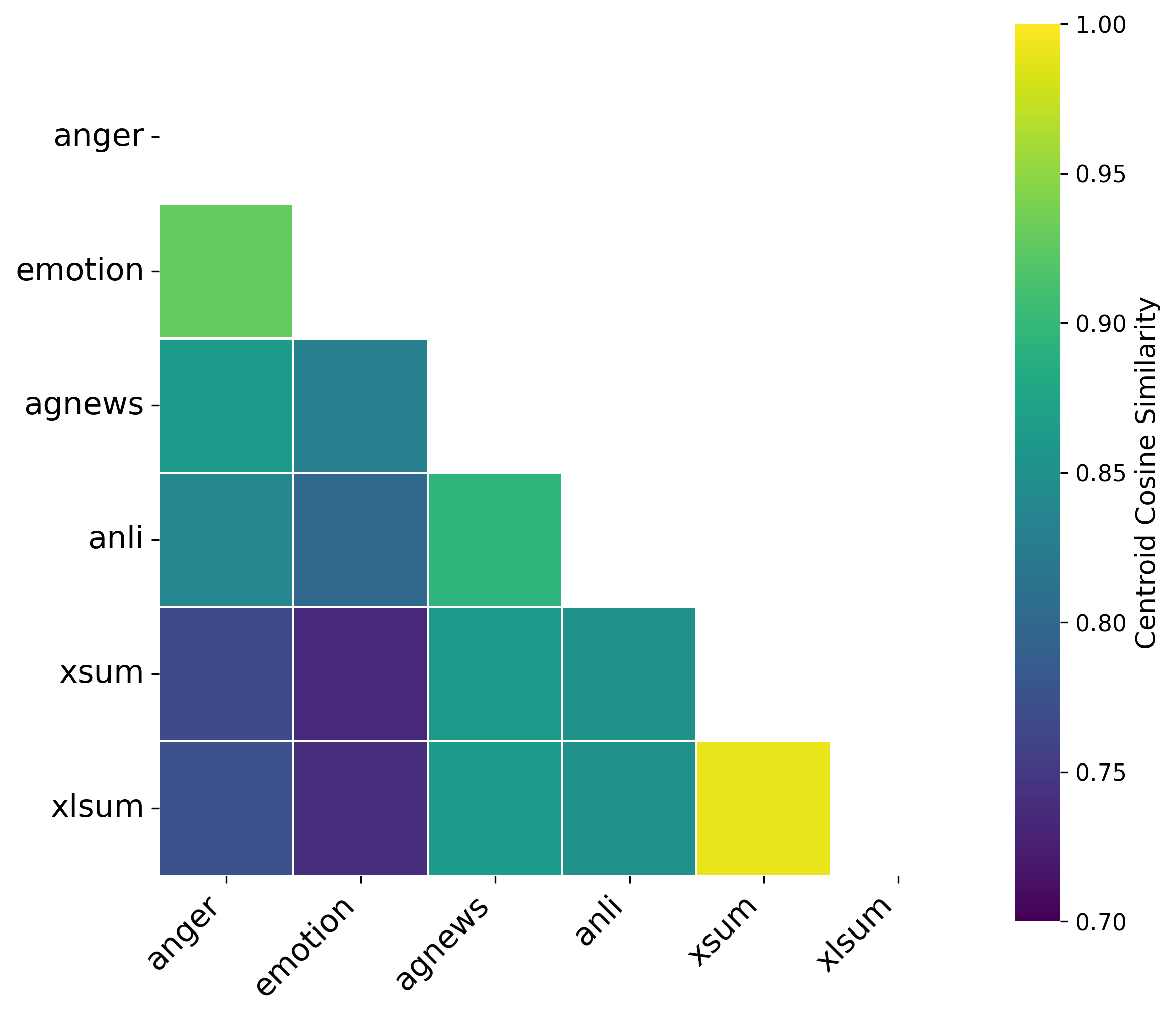}
    \caption{Centroid cosine similarity.}
    \label{fig:task-sim-centroid}
  \end{subfigure}
  \hfill
  \begin{subfigure}[t]{0.49\linewidth}
    \centering
    \includegraphics[width=\linewidth]{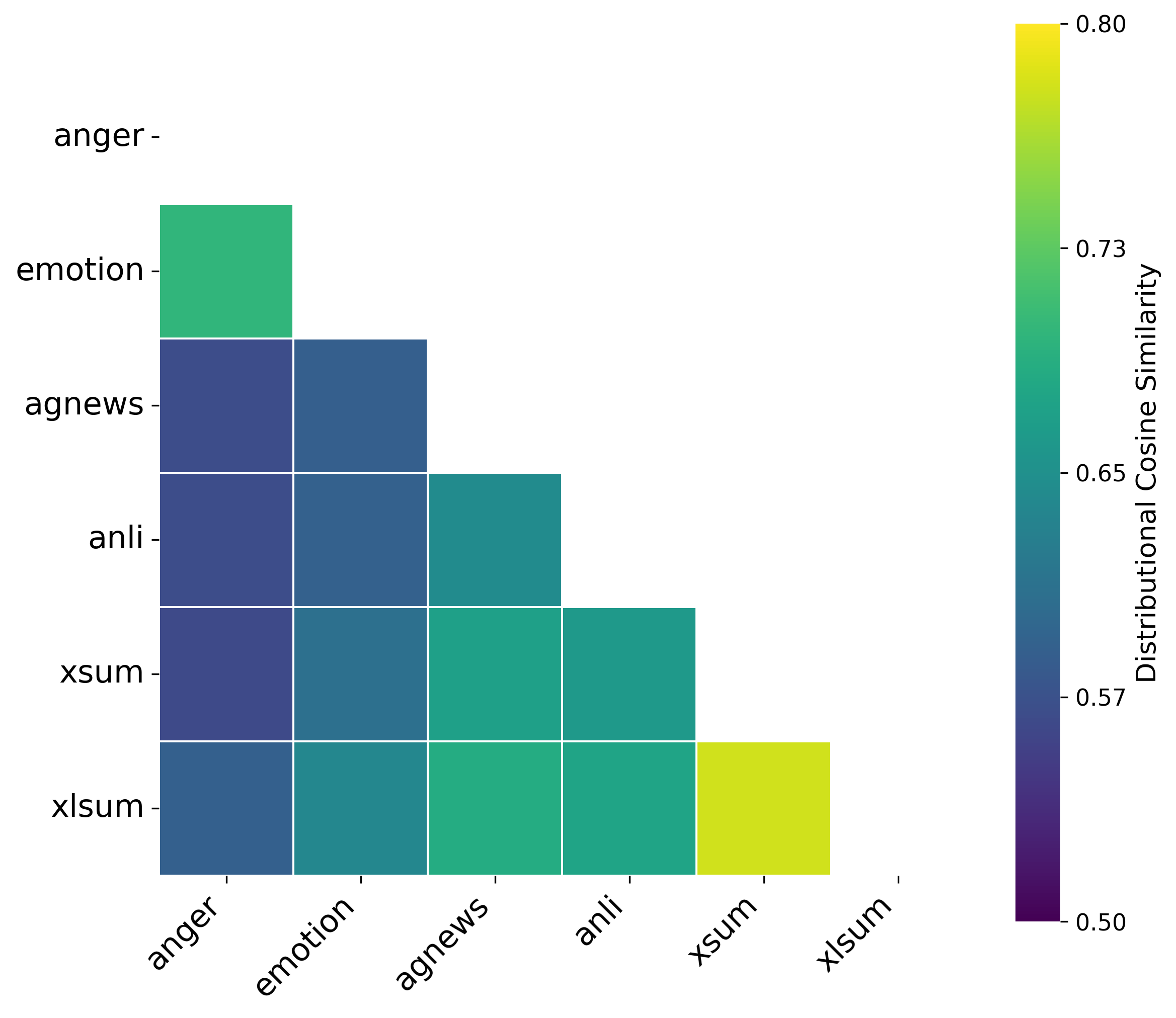}
    \caption{Distributional cosine similarity.}
    \label{fig:task-sim-dist}
  \end{subfigure}

  \caption{Cross-task similarity in representation space.}
  \label{fig:task-sim}
\end{figure*}

All datasets are in English. Beyond individual task details, we characterize similarity across tasks in representation space. Specifically, we compute cosine similarity at both centroid and distributional levels based on representations from a fixed encoder. As shown in Figure~\ref{fig:task-sim}, the benchmark exhibits substantial heterogeneity in cross-task similarity, xsum and xlsum are highly similar, anger and emotion show moderate similarity, while most other task pairs are well separated. Notably, we do not compute any predefined task similarity before the task sequence, nor use it to make any task-level decisions beyond the current stage. At each stage, only the current task data and the previous task prototype are available.

\subsection{Baseline Details}

\paragraph{\textbf{Sequence}.}
Sequence is the simplest CL baseline. A single PEFT module is trained sequentially, using only the current task’s data at each stage, following the predefined task order. After finishing task $t$, the same PEFT parameters are reused as initialization for task $t{+}1$, without any explicit anti-forgetting mechanism. 

\paragraph{\textbf{Replay}.}
Replay augments sequential training with a memory buffer that stores samples from previous tasks. At each stage, training batches are formed by mixing current data with replay samples. The PEFT module is still a single adapter updated sequentially, but replay provides a direct rehearsal signal that helps retain past knowledge. The buffer is updated after each stage by adding samples from the current task.

\paragraph{\textbf{EWC}.}
EWC adds a quadratic penalty that discourages changing parameters deemed important for previous tasks during training on a new task \cite{doi:10.1073/pnas.1611835114}. In our PEFT setting, the EWC penalty is applied only to trainable PEFT parameters, while the backbone remains frozen.

Concretely, after finishing task $t$, we estimate a diagonal Fisher approximation $F_{t,i}$ for each trainable PEFT parameter $\theta_i$ based on squared loss gradients on task $t$. Let $\theta_s^{*}$ denote the learned parameter vector after task $s$, and $\theta^{*}_{s,i}$ its $i$-th entry. We adopt an online accumulated variant that maintains:
\begin{equation}
\label{eq:ewc_running}
P_i=\sum_{s\le t}F_{s,i},
\ \ 
\mu_i=\frac{\sum_{s\le t}F_{s,i}\theta^{*}_{s,i}}{P_i}.
\end{equation}

When learning task $t{+}1$, EWC adds a quadratic penalty:
\begin{equation}
\label{eq:ewc_penalty}
\mathcal{L}_{\mathrm{EWC}}(\theta)
= \frac{\lambda_{\mathrm{EWC}}}{2}\sum_i P_i\left(\theta_i-\mu_i\right)^2,
\end{equation}
which discourages changing parameters that are important for previously learned tasks.

\paragraph{\textbf{MoE}.}
Our MoE baseline maintains multiple PEFT experts and uses a learned router to softly combine experts for each input during sequential training. Instead of a single PEFT module, each target module is equipped with $N_e$ PEFT experts, where expert $e$ provides a parameterized update $\Delta_e(\cdot)$. Let $x$ denote the layer input and $W_{\mathrm{base}}$ the frozen backbone weight. The MoE layer output is:
\begin{equation}
\label{eq:moe_peft}
y = x W_{\mathrm{base}} + \sum_{e=1}^{N_e} \pi_e(x)\,\Delta_e(x),
\end{equation}
where $\pi(x)\in\mathbb{R}^{E}$ is the routing distribution produced by a shared router.

In our implementation, a shared router computes the routing distribution $\pi(x)$ from a pooled layer input. Specifically, it first averages token representations into a single vector, then feeds it into a lightweight two-layer MLP with a softmax output. During training, we freeze the backbone and update only the router and the PEFT expert parameters.

\paragraph{\textbf{Joint}.}
Joint trains a single PEFT module on the union of data from all tasks. It does not follow the continual learning constraint because it assumes that all task data are available for training.

\paragraph{\textbf{Separate}.}
Separate trains an independent PEFT module for each task, without parameter sharing across tasks. At evaluation time, it selects the corresponding task-specific module using the given task identity. 

\subsection{Metric Details}
\label{app:metrics}
Let $T$ be number of tasks in a sequence. Let $a_{i,j}$ denote performance on test set of task $j$ after completing training on task $i$, where $i\in\{1,\ldots,T\}$ and $j\in\{1,\ldots,T\}$. For classification tasks, $a_{i,j}$ is accuracy, and for generation tasks, $a_{i,j}$ is ROUGE-L. Let $a_{0,j}$ denote performance on task $j$ evaluated at initialization, before any training is performed. Let $a^{\mathrm{sep}}_{j}$ denote performance of a model trained on task $j$ only in a single-task setting.

\paragraph{\textbf{Average Performance (AP)}.}
AP measures overall ability after learning full sequence, computed as mean final performance across tasks:
\begin{equation}
\label{eq:aa}
\mathrm{AP}
=
\frac{1}{T}\sum_{j=1}^{T} a_{T,j}.
\end{equation}

\paragraph{\textbf{Forgetting (FGT)}.}
Forgetting quantifies worst-case performance drop on each past task over training. For each task $j\le T-1$, define:
\begin{equation}
\label{eq:best_hist}
a^{\max}_j = \max_{i=j,\ldots,T} a_{i,j},
\end{equation}

\begin{equation}
\label{eq:forget_task}
F_j = a^{\max}_j - a_{T,j},
\ \ j=1,\ldots,T-1,
\end{equation}
where $a^{\max}_j$ is best performance achieved on task $j$ after it is first learned. Mean forgetting over the first $T-1$ tasks is:
\begin{equation}
\label{eq:forget_mean}
\mathrm{Forgetting}
=
\frac{1}{T-1}\sum_{j=1}^{T-1} F_j .
\end{equation}

\paragraph{\textbf{Forward Transfer (FWT)}.}
FWT measures benefit of prior learning on a new task. Two variants are reported.

(1) FWT-1 (zero-shot reference):
\begin{equation}
\label{eq:fwt1}
\mathrm{FWT\mbox{-}1}
=
\frac{1}{T-1}\sum_{j=2}^{T}\big(a_{j-1,j}-a_{0,j}\big),
\end{equation}
where $a_{j-1,j}$ evaluates model after task $j-1$ on task $j$ without training on task $j$.

(2) FWT-2 (single-task reference):
\begin{equation}
\label{eq:fwt2}
\mathrm{FWT\mbox{-}2}
=
\frac{1}{T}\sum_{j=1}^{T}\big(a_{j,j}-a^{\mathrm{sep}}_{j}\big).
\end{equation}

\paragraph{\textbf{Backward Transfer (BWT)}.}
BWT measures effect of later learning on earlier tasks by comparing final performance on each past task to performance right after learning that task:
\begin{equation}
\label{eq:bwt}
\mathrm{BWT}
=
\frac{1}{T-1}\sum_{j=1}^{T-1}\big(a_{T,j}-a_{j,j}\big).
\end{equation}

\begin{table*}[h]
\centering
\caption{Discrete recovery statistics under embedding-level PGD-1 perturbations with $\epsilon{=}0.1$.
All values are multiplied by $100$ for readability.}
\label{tab:discrete-recovery}
\begin{tabular}{l|c|c|c|c}
\toprule
\textbf{Task} & \textbf{Overall Shift} & \textbf{Per-Token Shift} & \textbf{Token Change Rate} & \textbf{Min Margin} \\
\midrule
anger   & 4.72 & 0.75 & 0.00 & 7.78 \\
emotion & 3.81 & 0.67 & 0.00 & 13.33 \\
agnews  & 4.47 & 0.46 & 0.00 & 7.78 \\
anli    & 4.72 & 0.30 & 0.00 & 7.47 \\
xsum    & 5.10 & 0.13 & 0.00 & 7.28 \\
xlsum   & 5.72 & 0.13 & 0.00 & 7.28 \\
\bottomrule
\end{tabular}
\end{table*}

\subsection{Training Details}
\label{app:training details}
All experiments run on a single NVIDIA A100 40GB GPU. We use Llama~3.2~3B Instruct\footnote{\url{https://huggingface.co/meta-llama/Llama-3.2-3B-Instruct}} as the frozen backbone and adopt LoRA \cite{hu2022lora} as the PEFT module.

For non-MoE baselines, i.e., Sequence, Replay, EWC, Joint, and Separate, LoRA uses rank $r{=}32$, scaling factor $\alpha{=}64$, and $\text{dropout}{=}0.1$. Target modules are \texttt{q\_proj}, \texttt{k\_proj}, \texttt{v\_proj}, \texttt{o\_proj}, \texttt{gate\_proj}, \texttt{up\_proj}, and \texttt{down\_proj}. For MoE, we use $N_e{=}6$ LoRA experts. Each expert uses rank $r{=}32$, $\alpha{=}64$, and $\text{dropout}{=}0.1$, and MoE-LoRA is applied only to \texttt{q\_proj}, \texttt{k\_proj}, \texttt{v\_proj}, and \texttt{o\_proj}.

For Replay, after finishing each task, we randomly sample $0.02$ of its training samples and store them for replay in subsequent stages.

For Intra-Smooth, we sample a subset $\mathcal{B}_{\mathrm{intra}}$ from each mini-batch with ratio $r_{\mathrm{intra}}{=}0.25$. We set $\epsilon_{\mathrm{intra}}{=}0.1$, $k_{\mathrm{intra}}{=}1$, and $\lambda_{\mathrm{intra}}{=}0.1$. For Proto-Clip, we use momentum coefficient $m{=}0.9$ and threshold $\tau_{\mathrm{hi}}{=}0.8$, with $\lambda_{\mathrm{clip}}{=}0.5$. 
For Inter-Align, we sample a subset $\mathcal{B}_{\mathrm{inter}}$ from each mini-batch with ratio $r_{\mathrm{inter}}{=}0.125$. We set $\epsilon_{\mathrm{inter}}{=}0.03$, $k_{\mathrm{inter}}{=}1$, and $\lambda_{\mathrm{inter}}{=}0.05$.

For classification tasks, the micro-batch size is 16 with gradient accumulation of 8, yielding an effective batch size of 128. For generation tasks, the micro-batch size is 4 with gradient accumulation of 32. We use a learning rate of 3e-5 with a warmup ratio of 0.05 and weight decay 0.1.

We set maximum input length to 512 tokens and maximum output length to 64 tokens. For each stage, we train for 160 steps if current task is a classification task, and for 20 steps if it is a generation task.

For generation evaluation, we use beam search with 4 beams, prevent 3-gram repetitions, and set the length penalty to 1.1. All reported results are averaged over three runs with different random seeds.

\section{Discrete Recovery from Continuous Perturbations}
\label{app:discovery}
To assess discrete visibility of embedding-level continuous perturbations, we perform a discrete recovery analysis. Given an input token sequence $x=(x_1,\ldots,x_L)$, its embedding representation is $E(x)\in\mathbb{R}^{L\times d_e}$. We apply a continuous perturbation $\Delta\in\mathbb{R}^{L\times d_e}$ to obtain the perturbed embeddings $E(x)+\Delta$. To facilitate analysis, we map each perturbed embedding vector $(E(x)+\Delta)_i$ back to a proxy discrete token by nearest-neighbor retrieval in the vocabulary embedding space using cosine similarity. This recovery is used only to characterize how continuous perturbations translate into discrete token changes, and should not be interpreted as explicit discrete substitution attacks.

We report four metrics: (1) \textbf{Overall Shift}, $\lVert \Delta\rVert_F$ averaged across samples; (2) \textbf{Per-token Shift}, $\lVert \Delta_i\rVert_2$ averaged across token positions within each sample, and then across samples; (3) \textbf{Token Change Rate}, the fraction of token positions where the nearest-neighbor recovered token differs from the original token, i.e., the proportion of positions whose discrete token changes after recovery; and (4) \textbf{Min Margin}, the minimum cosine similarity gap $s_{\mathrm{orig}}-s_{\mathrm{alt}}$, where $s_{\mathrm{orig}}$ is the cosine similarity between the perturbed embedding and the original token embedding, and $s_{\mathrm{alt}}$ is the highest cosine similarity to any alternative token. Smaller values indicate the recovered discrete token is closer to changing.

Table~\ref{tab:discrete-recovery} reports results across all six tasks under PGD-1 with $\epsilon{=}0.1$. Both Overall Shift and Per-Token Shift are consistently non-zero across tasks, indicating a stable perturbation size in embedding space. In contrast, Token Change Rate is $0$ for all tasks and Min margin remains positive, suggesting that nearest-neighbor recovery yields the same discrete tokens as the original input. These observations indicate that embedding-level perturbations can substantially modify continuous representations while leaving the recovered discrete tokens unchanged, which explains why the perturbations are largely imperceptible at the token level.

\section{Additional Order Results}
\label{app:additional order}

In addition to default task order, we also evaluate baselines under an alternative order: emotion $\rightarrow$ agnews $\rightarrow$ anger $\rightarrow$ xsum $\rightarrow$ xlsum $\rightarrow$ anli. Results are reported in Table~\ref{tab:additional-order}. For the alternative task order, we directly reuse the hyperparameters from the default task order without further tuning. Overall, trends are consistent with Table~\ref{tab:main_standard}, indicating that our modules yields stable improvements across different task orders.

\begin{table*}[!t]
\centering
\caption{
Results of additional task order under Standard and PGD settings.
$\uparrow$ indicates higher is better, while $\downarrow$ indicates lower is better.
+ S, + C, and + A denote augmenting a baseline with Intra-Smooth, Proto-Clip, and Inter-Align, respectively.
The best and second-best results are highlighted in \textbf{bold} and \uline{underlined}, respectively.
}
\label{tab:additional-order}

\resizebox{\textwidth}{!}{
\begin{tabular}{l|ccccc|ccccc}
\toprule
 & \multicolumn{5}{c|}{\textbf{Standard}} & \multicolumn{5}{c}{\textbf{PGD (Input)}}\\
 & \textbf{AP} & \textbf{FGT} & \textbf{FWT-1} & \textbf{FWT-2} & \textbf{BWT}
 & \textbf{AP} & \textbf{FGT} & \textbf{FWT-1} & \textbf{FWT-2} & \textbf{BWT} \\
\textbf{Model} & $(\uparrow)$ & $(\downarrow)$ & $(\uparrow)$ & $(\uparrow)$ & $(\uparrow)$
               & $(\uparrow)$ & $(\downarrow)$ & $(\uparrow)$ & $(\uparrow)$ & $(\uparrow)$ \\
\midrule[0.9pt]
\textbf{Sequence}
& 53.15 & 3.70 & 0.24 & 0.42 & -2.98
& 32.34 & 8.47 & 2.21 & -0.04 & -7.89 \\
+ S (Smooth)
& \uline{55.43} & \uline{2.30} & -0.64 & 1.61 & \uline{-1.66}
& \uline{39.63} & \uline{4.60} & \textbf{4.49} & \textbf{4.38} & \uline{-4.45} \\
+ C (Clip)
& 54.62 & 3.41 & \uline{0.56} & \textbf{2.25} & -3.41
& 30.47 & 10.69 & 1.64 & -0.42 & -9.69 \\
+ A (Align)
& 53.94 & 3.97 & \textbf{0.77} & 1.73 & -3.60
& 30.17 & 11.07 & 2.21 & -0.73 & -9.67 \\
+ S + C + A
& \textbf{56.09} & \textbf{1.63} & -0.87 & \uline{1.80} & \textbf{-1.11}
& \textbf{42.12} & \textbf{1.15} & \uline{2.54} & \uline{2.96} & \textbf{0.24} \\
\bottomrule
\end{tabular}
}
\end{table*}

\section{Robustness Evaluation}
\subsection{Cross-Attack Evaluation}
\label{app:cross}
We evaluate different baselines under five attack settings: PGD-5, PGD-10, FGSM, Rand, and explicit discrete perturbations. For PGD-5, PGD-10, and Rand, perturbations are constrained in $\ell_2$ norm with budget $\varepsilon{=}0.1$ in embedding space. For FGSM, perturbations are constrained in $\ell_\infty$ norm with budget $\varepsilon{=}\text{5e-4}$ in embedding space. For discrete perturbations, we apply TextAttack\footnote{\url{https://github.com/QData/TextAttack}} to perturb 10\% of words for classification tasks and 5\% of words for generation tasks via synonym substitutions and typographical perturbations.

The averaged results across five attacks are reported in Table~\ref{tab:cross_attack_avg_5}. Intra-Smooth is the key contributor to robustness improvements. Inter-Align mainly helps mitigate forgetting and improves BWT. Combining all three modules further strengthens robustness beyond Intra-Smooth alone, achieving the best overall performance across all metrics.

\begin{table*}[t]
\centering
\caption{Cross-attack results averaged over five attacks (PGD-5, PGD-10, FGSM, Rand, and discrete perturbations).}
\label{tab:cross_attack_avg_5}
\begin{tabular}{l|ccccc}
\toprule
\textbf{Model} & \textbf{AP$(\uparrow)$} & \textbf{FGT$(\downarrow)$} & \textbf{FWT-1$(\uparrow)$} & \textbf{FWT-2$(\uparrow)$} & \textbf{BWT$(\uparrow)$} \\
\midrule
\textbf{Sequence} & 37.47 & 8.78 & 2.20 & 1.61 & -7.52 \\
+ S (Smooth) & 43.23 & 4.04 & \textbf{3.72} & 2.50 & -1.68 \\
+ C (Clip) & 36.24 & 9.76 & 1.59 & 0.67 & -7.87 \\
+ A (Align) & 39.33 & 5.14 & 1.67 & -0.38 & -3.46 \\
+ S + C + A & \textbf{44.22} & \textbf{2.76} & 3.48 & \textbf{2.90} & \textbf{-0.40} \\
\midrule
\textbf{Replay} & 40.85 & 3.89 & 2.38 & 0.23 & -1.81 \\
+ S (Smooth) & \textbf{45.75} & \textbf{2.61} & \textbf{5.15} & \textbf{4.39} & \textbf{-0.93} \\
+ C (Clip) & 41.39 & 4.13 & 1.23 & 0.99 & -2.08 \\
+ A (Align) & 41.60 & 3.66 & 1.22 & 0.97 & -1.80 \\
\midrule
\textbf{EWC} & 36.61 & 10.12 & 1.60 & 1.21 & -8.08 \\
+ S (Smooth) & \textbf{43.31} & \textbf{4.18} & \textbf{3.55} & \textbf{2.71} & \textbf{-1.83} \\
+ C (Clip) & 38.02 & 7.50 & 1.64 & 0.39 & -5.40 \\
+ A (Align) & 36.76 & 8.83 & 2.17 & 0.29 & -6.79 \\
\midrule
\textbf{MoE} & 34.45 & 12.43 & 0.45  & \textbf{1.30}  & -10.78 \\
+ S (Smooth)& \textbf{35.73} & \textbf{9.25} & \textbf{1.41}  & -0.13  & \textbf{-7.92} \\
+ C (Clip)& 34.20 & 11.37 & 0.64  & 0.54  & -10.17  \\
+ A (Align)& 35.23 & 10.48 &  1.40 & 0.71  & -9.13  \\
\bottomrule
\end{tabular}
\end{table*}

\subsection{Robustness under PGD}
\label{app:robust pgd}
\begin{figure*}[t]
	\centering
	\includegraphics[width=0.7\linewidth]{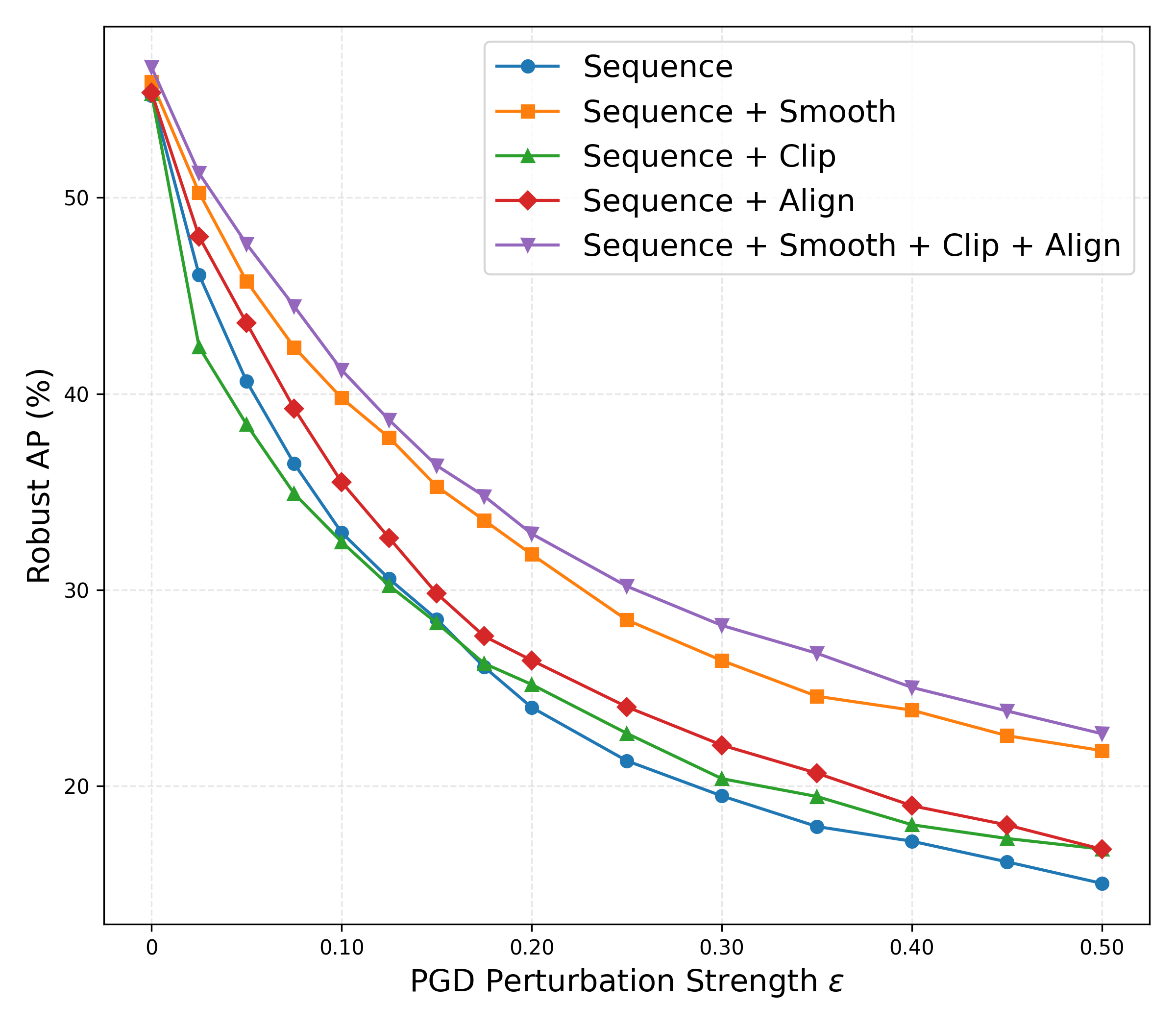}
	\caption{Robustness curve under PGD.
}
	\label{fig:ch6_a}
\end{figure*}

As shown in Figure~\ref{fig:ch6_a}, robust AP decreases monotonically as perturbation strength $\epsilon$ increases. Intra-Smooth yields consistent gains, since its perturbation form is more aligned with PGD evaluation. In contrast, mechanisms of Proto-Clip and Inter-Align are not fully matched to this evaluation setting, yet combining three modules still maintains higher robustness, indicating complementary gains among modules.

\section{Geometric Metrics for Inter-Align}
\label{app:geometric metrics}

\subsection{Metric Definitions}
Let $E(\cdot)$ denote the encoder that maps an input sample to a representation vector. We use $\ell_2$ normalized representations $\mathbf{z}=E(x)/\|E(x)\|_2$ and compute cosine similarity and cosine distance accordingly. For each adjacent task pair, we denote the previous task by $\mathrm{p}$ and the current task by $\mathrm{c}$. Let $\mathcal{Z}_{\mathrm{p}}=\{\mathbf{z}^{\mathrm{p}}_i\}_{i=1}^{n_{\mathrm{p}}}$ and $\mathcal{Z}_{\mathrm{c}}=\{\mathbf{z}^{\mathrm{c}}_j\}_{j=1}^{n_{\mathrm{c}}}$ be the sets of normalized representations from previous current tasks, respectively. We use $S(\mathbf{u},\mathbf{v})$ for cosine similarity and $1-S(\mathbf{u},\mathbf{v})$ for cosine distance.

\paragraph{\textbf{Centroid Cosine Distance}.}
We represent each task by a centroid prototype in representation space. Specifically, we compute the mean representation within each task and then apply $\ell_2$ normalization:
\begin{equation}
\bar{\mathbf{z}}_{\mathrm{p}}=\frac{1}{n_{\mathrm{p}}}\sum_{i=1}^{n_{\mathrm{p}}}\mathbf{z}^{\mathrm{p}}_i,\ \ 
\bar{\mathbf{z}}_{\mathrm{c}}=\frac{1}{n_{\mathrm{c}}}\sum_{j=1}^{n_{\mathrm{c}}}\mathbf{z}^{\mathrm{c}}_j,
\end{equation}
\begin{equation}
\hat{\mathbf{z}}_{\mathrm{p}}=\frac{\bar{\mathbf{z}}_{\mathrm{p}}}{\|\bar{\mathbf{z}}_{\mathrm{p}}\|_2},\ \ 
\hat{\mathbf{z}}_{\mathrm{c}}=\frac{\bar{\mathbf{z}}_{\mathrm{c}}}{\|\bar{\mathbf{z}}_{\mathrm{c}}\|_2}.
\end{equation}

Centroid cosine distance is defined as:
\begin{equation}
d_{\mathrm{cent}}=1-S\!\left(\hat{\mathbf{z}}_{\mathrm{p}},\hat{\mathbf{z}}_{\mathrm{c}}\right).
\end{equation}

This metric captures global gap between the mean representation prototypes of previous and current tasks. A smaller value indicates closer centroids. If this value decreases after adding Inter-Align, it suggests that Inter-Align improves cross-task geometric alignment by bringing adjacent task centroids closer and reducing global representational shift.

\paragraph{\textbf{Mean Pairwise Cosine Distance}.}
Centroid Cosine Distance reflects only the difference between task centroids, so it may miss finer distributional differences. We therefore compute Mean Pairwise Cosine Distance by averaging cosine distances over cross-task pairs:
\begin{equation}
\begin{aligned}
d_{\mathrm{pair}} &= \mathbb{E}_{(\mathbf{z}_{\mathrm{p}},\mathbf{z}_{\mathrm{c}})}
\!\left[1-S(\mathbf{z}_{\mathrm{p}},\mathbf{z}_{\mathrm{c}})\right], \\
&\text{where } \mathbf{z}_{\mathrm{p}} \in \mathcal{Z}_{\mathrm{p}},\ \text{and }
\mathbf{z}_{\mathrm{c}} \in \mathcal{Z}_{\mathrm{c}}.
\end{aligned}
\end{equation}

This metric measures distribution level gap between previous and current tasks by averaging cosine distances over cross-task sample pairs. Lower values mean the two tasks are closer on average at sample level. Therefore, a decrease after adding Inter-Align suggests improved cross-task geometric alignment by narrowing the distribution level gap, not only the centroid gap.

\paragraph{\textbf{Directional Prototype Alignment}.}
While the previous two metrics quantify centroid and distribution gaps, they do not indicate whether current task representations are aligned toward the previous task prototype. To capture this, we treat the previous task centroid $\hat{\mathbf{z}}_{\mathrm{p}}$ as a prototype and measure how strongly current task representations align with it:
\begin{equation}
s_{\mathrm{c\rightarrow p}}=\frac{1}{n_{\mathrm{c}}}\sum_{j=1}^{n_{\mathrm{c}}}S\!\left(\mathbf{z}^{\mathrm{c}}_j,\hat{\mathbf{z}}_{\mathrm{p}}\right).
\end{equation}

This metric measures alignment strength from current task representations to previous task prototype. Larger $s_{\mathrm{c\rightarrow p}}$ indicates stronger alignment, consistent with the intended effect of Inter-Align.
\subsection{Results}
For each adjacent task pair, Table~\ref{tab:inter_smooth_deltas} reports the absolute differences of three geometric metrics, computed as the difference between sequence training with Inter-Align enabled and sequence training without Inter-Align, using representations extracted after finishing training on each current task. We denote these differences by $\Delta d_{\mathrm{cent}}$, $\Delta d_{\mathrm{pair}}$, and $\Delta s_{\mathrm{c\rightarrow p}}$, respectively.

For $\Delta d_{\mathrm{cent}}$ and $\Delta d_{\mathrm{pair}}$, a negative value indicates improvement because it means a reduced geometric gap. For $\Delta s_{\mathrm{c\rightarrow p}}$, a positive value indicates improvement because it means current samples align more strongly with the previous task prototype in representation space.

Across adjacent task pairs, Inter-Align consistently reduces centroid gaps, since $\Delta d_{\mathrm{cent}}$ is negative for all pairs. For four of the five adjacent task pairs, it also reduces distribution level gaps and increases prototype alignment. These results support the claim that Inter-Align improves cross-task geometric alignment by narrowing representation gaps and pulling current task representations toward the previous task prototype.

For xsum$\rightarrow$xlsum, we observe a small inconsistency in that centroid gap is further reduced, whereas distribution level and alignment deltas take opposite signs compared with most adjacent task pairs. A plausible reason is that xsum and xlsum are highly similar, so centroid gap is already very small and Inter-Align mainly induces fine grained redistribution of representations, which can lead to small fluctuations at distribution level and in prototype alignment.

\begin{table}[t]
\caption{Absolute differences of geometric metrics for adjacent task pairs, reported in percentage points, where $\Delta(\cdot)=(\cdot)_{\text{Sequence+Inter-Align}}-(\cdot)_{\text{Sequence}}$.}
\label{tab:inter_smooth_deltas}
\centering
\begin{tabular}{lccc}
\toprule
\textbf{Task Pair} &
$\Delta d_{\mathrm{cent}}$ &
$\Delta d_{\mathrm{pair}}$ &
$\Delta s_{\mathrm{c\rightarrow p}}$ \\
\midrule
anger$\rightarrow$emotion  & -0.087 & -0.065 & +0.090 \\
emotion$\rightarrow$agnews & -0.020 & -0.168 & +0.162 \\
agnews$\rightarrow$anli    & -0.086 & -0.051 & +0.063 \\
anli$\rightarrow$xsum      & -0.694 & -1.087 & +0.964 \\
xsum$\rightarrow$xlsum     & -0.144 & +0.373 & -0.051 \\
\bottomrule
\end{tabular}%
\end{table}

\section{Additional Analysis of Inter-Align}
\label{app:inter sensitivity}
\subsection{Similarity Shift}
\begin{figure*}[t]
	\centering
	\includegraphics[width=0.7\linewidth]{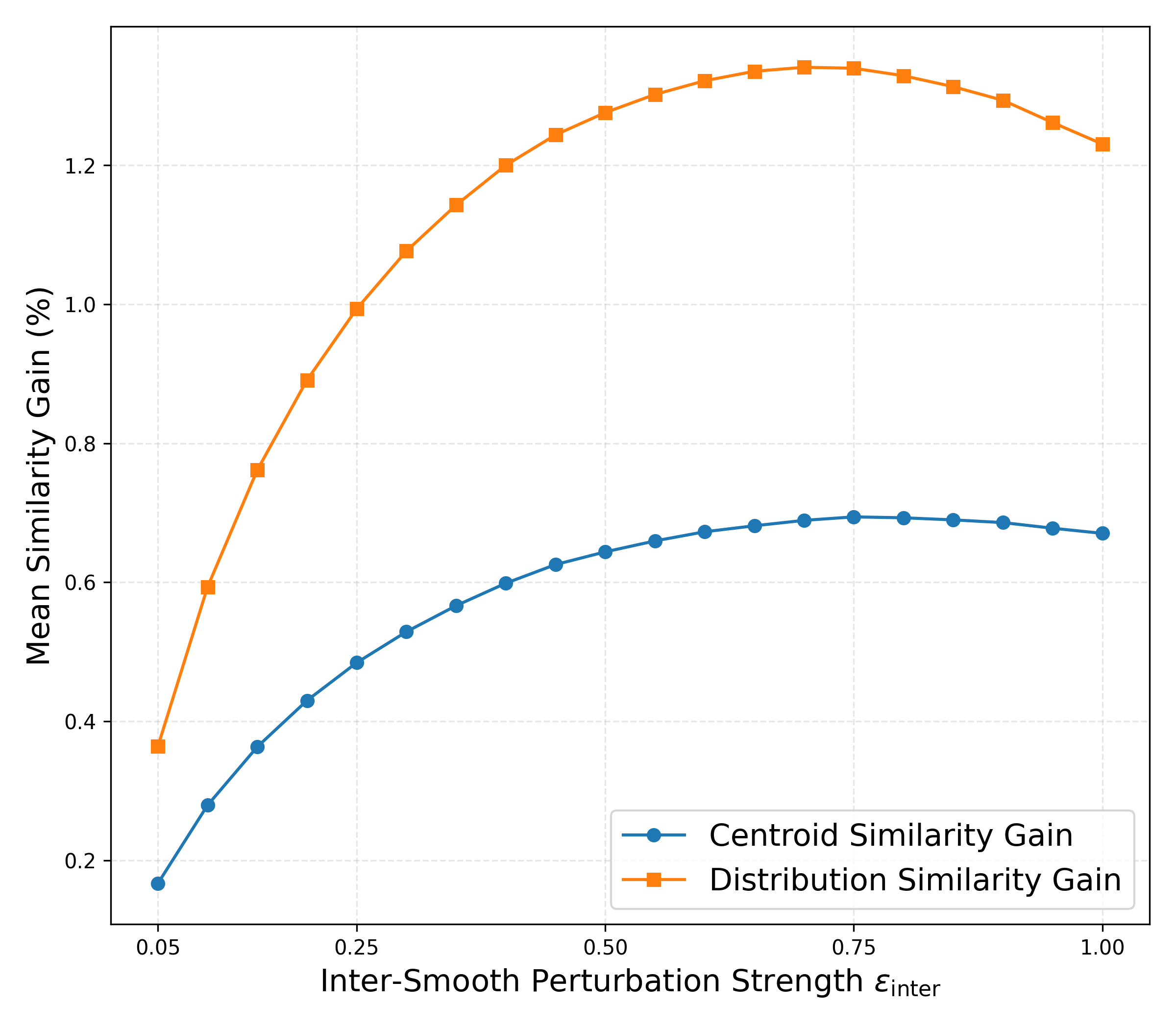}
	\caption{Similarity gain under varying $\epsilon_{\mathrm{inter}}$.
}
	\label{fig:ch6_b}
\end{figure*}

We study how Inter-Align shifts cross-task similarity in representation space of a fixed encoder by applying directional perturbations to current task samples toward previous task prototype. Figure~\ref{fig:ch6_b} shows that this perturbation increases cross-task similarity. As $\epsilon_{\mathrm{inter}}$ increases, similarity gain rises and gradually saturates for both centroid and mean pairwise (distribution) similarity, indicating that Inter-Align improves similarity in a tunable way.

\subsection{Inter-Align Sensitivity to Perturbation Strength}

\begin{table*}
\caption{Standard evaluation on Sequence baseline with Inter-Align under different perturbation strengths $\varepsilon_{\mathrm{inter}}$. We mainly focus on forgetting (\textbf{FGT}) and backward transfer (\textbf{BWT}). Finer-grained $\varepsilon_{\mathrm{inter}}$ values show the same trend and are omitted.}
\label{tab:inter-sensitivity}
\centering
\begin{tabular}{l|ccccc}
\toprule
\textbf{Setting} & \textbf{AP$(\uparrow)$} & \textbf{FGT$(\downarrow)$} & \textbf{FWT-1$(\uparrow)$} & \textbf{FWT-2$(\uparrow)$} & \textbf{BWT$(\uparrow)$} \\
\midrule[0.9pt]
\textbf{Sequence}  ($\varepsilon_{\mathrm{inter}}=0.00$) & 55.21 & 3.42 & 1.65 & 1.76 & -2.11 \\
+ Inter-Align ($\varepsilon_{\mathrm{inter}}=0.01$) & 55.00 & 1.80 & 1.79 & 0.72 & -1.11 \\
+ Inter-Align ($\varepsilon_{\mathrm{inter}}=0.02$) & 54.86 & 1.62 & 1.52 & 0.25 & -0.72 \\
+ Inter-Align ($\varepsilon_{\mathrm{inter}}=0.03$) & 55.36 & 2.20 & 0.28 & 1.23 & -1.29 \\
+ Inter-Align ($\varepsilon_{\mathrm{inter}}=0.04$) & 55.00 & 1.50 & 1.66 & 0.50 & -0.86 \\
+ Inter-Align ($\varepsilon_{\mathrm{inter}}=0.05$) & 54.91 & 2.11 & 1.76 & 0.79 & -1.31 \\
+ Inter-Align ($\varepsilon_{\mathrm{inter}}=0.10$) & 54.84 & 2.14 & 1.78 & 0.84 & -1.44 \\
+ Inter-Align ($\varepsilon_{\mathrm{inter}}=0.15$) & 54.87 & 1.88 & 1.90 & 0.56 & -1.08 \\
+ Inter-Align ($\varepsilon_{\mathrm{inter}}=0.20$) & 54.94 & 1.80 & 1.92 & 0.83 & -1.31 \\
+ Inter-Align ($\varepsilon_{\mathrm{inter}}=0.25$) & 54.98 & 1.80 & 1.67 & 0.65 & -1.05 \\
+ Inter-Align ($\varepsilon_{\mathrm{inter}}=0.30$) & 55.46 & 2.10 & 1.84 & 0.83 & -0.69 \\
\bottomrule
\end{tabular}
\end{table*}

We study the sensitivity of Inter-Align to perturbation strength $\varepsilon_{\mathrm{inter}}$ on Sequence baseline. As shown in Table~\ref{tab:inter-sensitivity}, Inter-Align consistently reduces forgetting across a wide range of $\varepsilon_{\mathrm{inter}}$ (e.g., FGT decreases from $3.42$ to around $1.50$--$2.20$) and improves backward transfer (BWT increases from $-2.11$ to as high as $-0.69$). These trends indicate that Inter-Align effectively stabilizes previously learned knowledge and alleviates catastrophic forgetting even when the alignment strength varies.

Meanwhile, AP is broadly stable across $\varepsilon_{\mathrm{inter}}$, with only minor decreases. This is expected because Inter-Align acts as a cross-task geometric regularizer by pulling a subset of current representations toward previous task prototype, and it is not designed to explicitly improve performance on current task. Consequently, it can impose a mild constraint that prioritizes preserving past knowledge, which may slightly reduce AP while keeping changes small and acceptable.

Overall, these results suggest that Inter-Align is robust to the choice of $\varepsilon_{\mathrm{inter}}$ and serves as an effective plug-in module for mitigating forgetting and improving backward transfer, with minimal impact on average performance.

\section{Computational Overhead Analysis}
\label{app:overhead}

We take one training step without any module enabled as the baseline computation (denoted as $1.0\times$), which consists of a standard forward and backward update. Below, we approximate additional computation by counting extra forward passes and gradient computations introduced by each module.

Intra-Smooth requires solving perturbation $\Delta_{\mathrm{intra}}^{\star}$ for a subset of samples. Specifically, to update $\Delta_{\mathrm{intra}}^{\star}$ (e.g., $k_{\mathrm{intra}}$ steps of PGD), we need extra forward passes and gradient computations to obtain $\nabla_{\Delta}$ on the subset. Since this process is applied only to a fraction $r_{\mathrm{intra}}$ of the batch, additional computation can be approximated as $r_{\mathrm{intra}} k_{\mathrm{intra}}$. Under default settings $r_{\mathrm{intra}}{=}0.25$ and $k_{\mathrm{intra}}{=}1$, Intra-Smooth incurs approximately +25\% (1.25$\times$) additional computation.

Inter-Align similarly solves directional perturbation $\Delta_{\mathrm{inter}}^{\star}$ on a fraction $r_{\mathrm{inter}}$ of the batch. Additional cost mainly comes from extra forward and gradient computations needed to update $\Delta_{\mathrm{inter}}^{\star}$, yielding an overhead of approximately $r_{\mathrm{inter}} k_{\mathrm{inter}}$. Under default settings $r_{\mathrm{inter}}{=}0.125$ and $k_{\mathrm{inter}}{=}1$, Inter-Align incurs approximately +12.5\% (1.125$\times$) additional computation.

Proto-Clip applies a similarity-based clipping regularizer by comparing representations from the baseline training step with the previous task prototype. In implementation, this module reuses the intermediate activations from the baseline step and is backpropagated jointly with the main loss, without requiring an extra model forward or backward pass. Therefore, its additional cost is dominated by lightweight similarity and clipping operations and is negligible in practice.

\section{Additional Performance Analysis}
\label{app:stagewise and tradeoff}

\subsection{Per-Task Stagewise Performance Curves}
\begin{figure*}[t]
  \centering

  \begin{subfigure}[t]{0.48\textwidth}
    \centering
    \includegraphics[width=\linewidth]{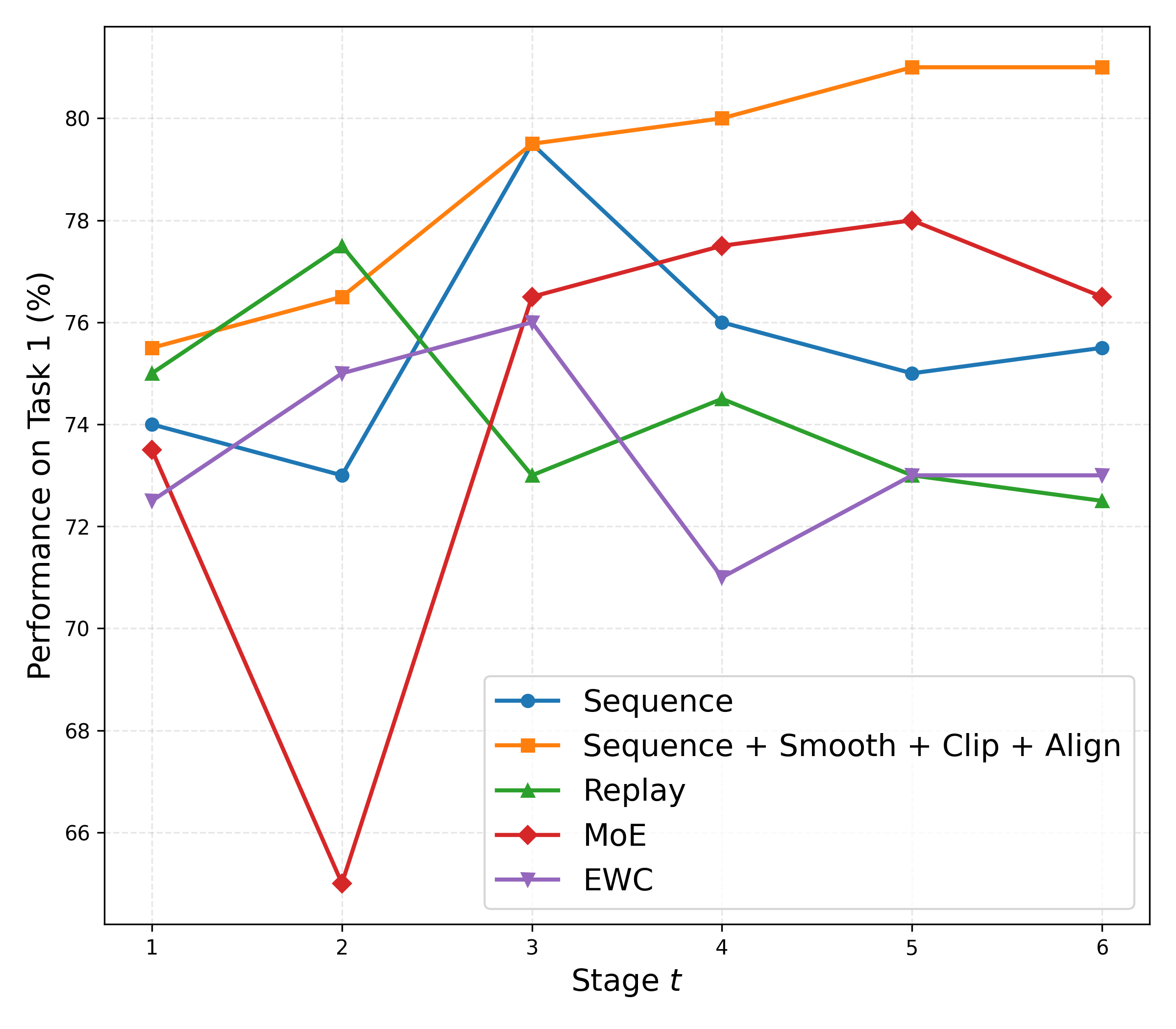}
    \caption{Task 1 (anger)}
    \label{fig:app_task1_curve}
  \end{subfigure}
  \hfill
  \begin{subfigure}[t]{0.48\textwidth}
    \centering
    \includegraphics[width=\linewidth]{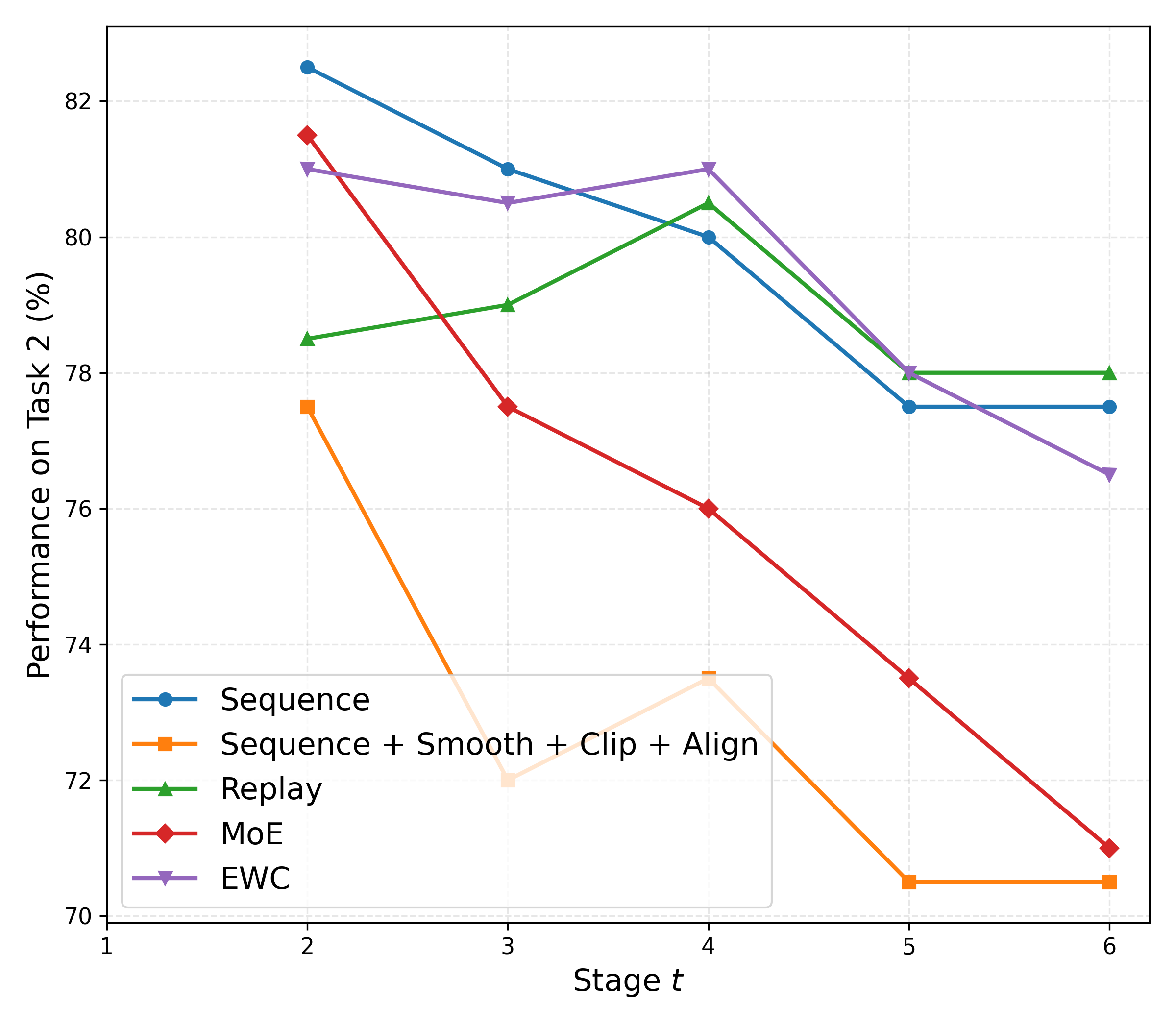}
    \caption{Task 2 (emotion)}
    \label{fig:app_task2_curve}
  \end{subfigure}

  \vspace{0.6em}

  \begin{subfigure}[t]{0.48\textwidth}
    \centering
    \includegraphics[width=\linewidth]{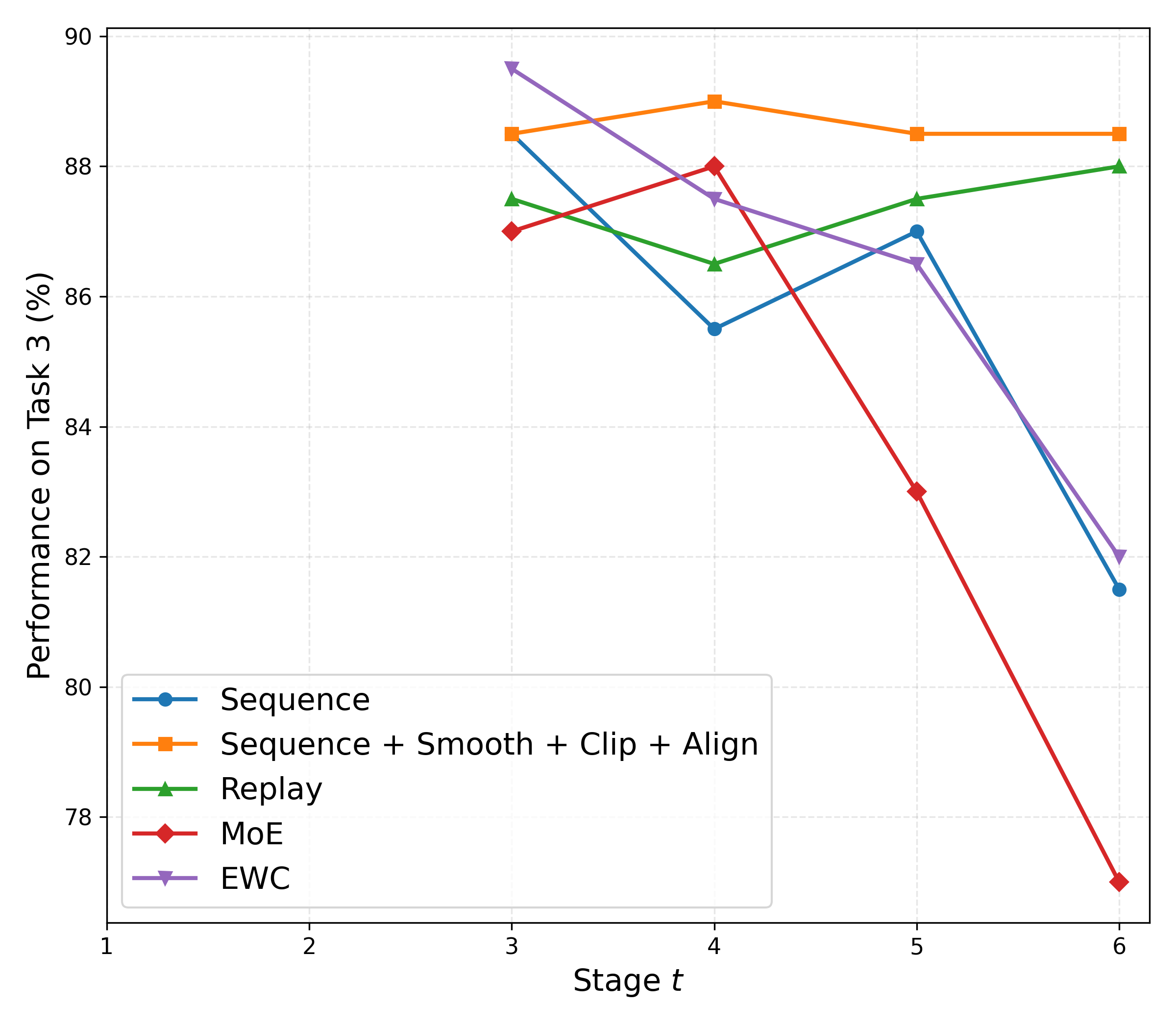}
    \caption{Task 3 (agnews)}
    \label{fig:app_task3_curve}
  \end{subfigure}
  \hfill
  \begin{subfigure}[t]{0.48\textwidth}
    \centering
    \includegraphics[width=\linewidth]{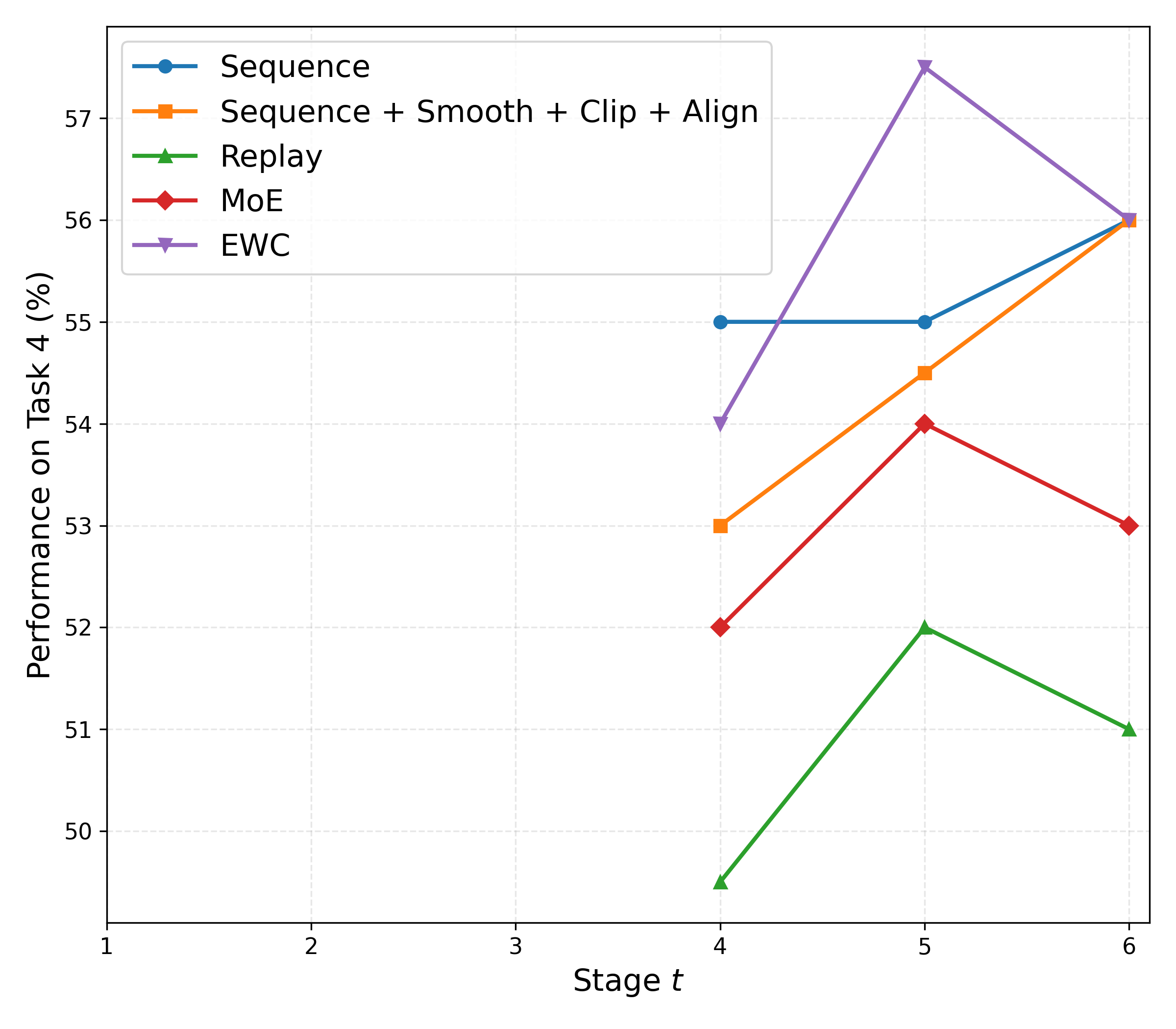}
    \caption{Task 4 (anli)}
    \label{fig:app_task4_curve}
  \end{subfigure}

  \vspace{0.6em}

  \begin{subfigure}[t]{0.48\textwidth}
    \centering
    \includegraphics[width=\linewidth]{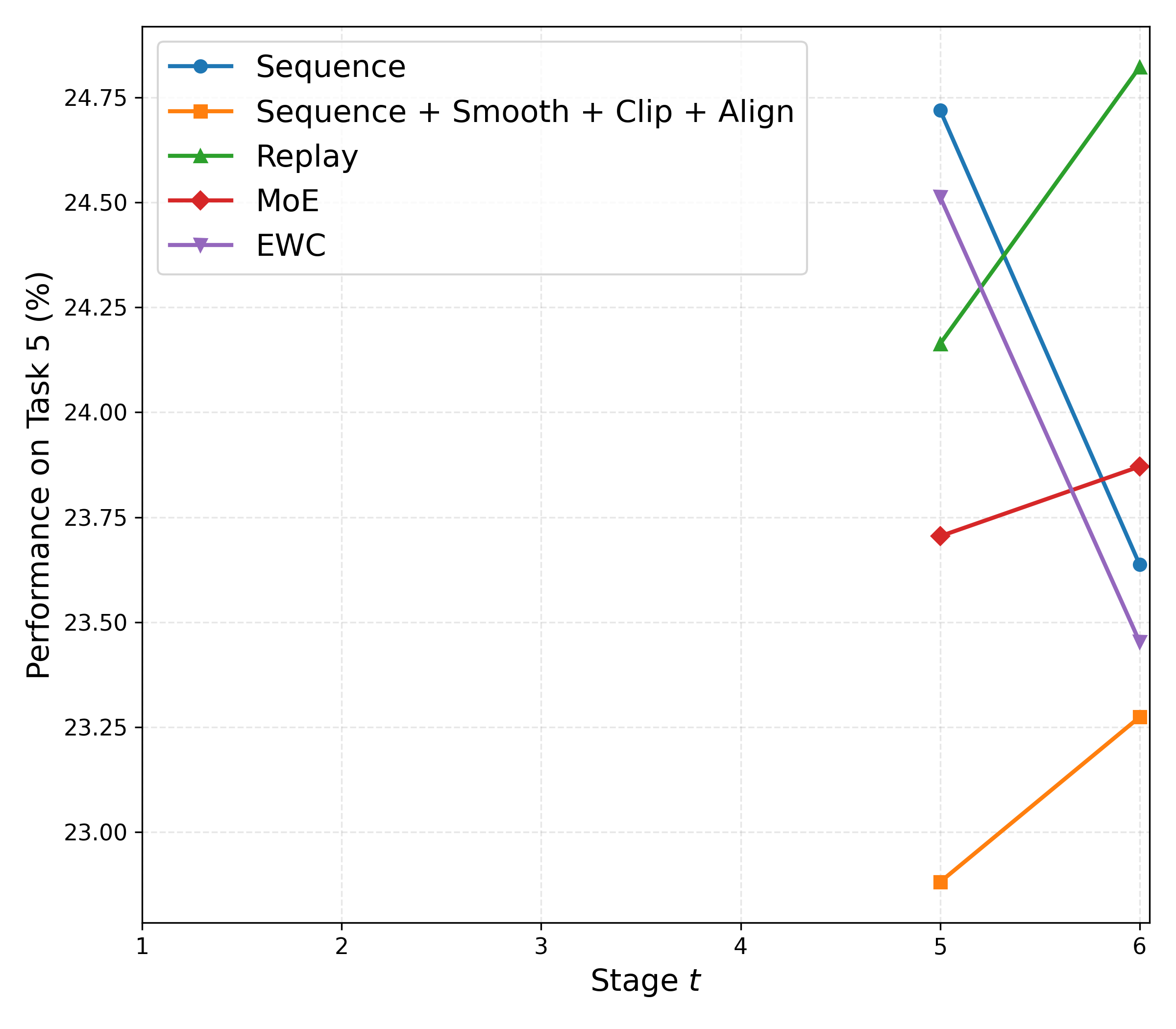}
    \caption{Task 5 (xsum)}
    \label{fig:app_task5_curve}
  \end{subfigure}
  \hfill
  \begin{subfigure}[t]{0.48\textwidth}
    \centering
    \includegraphics[width=\linewidth]{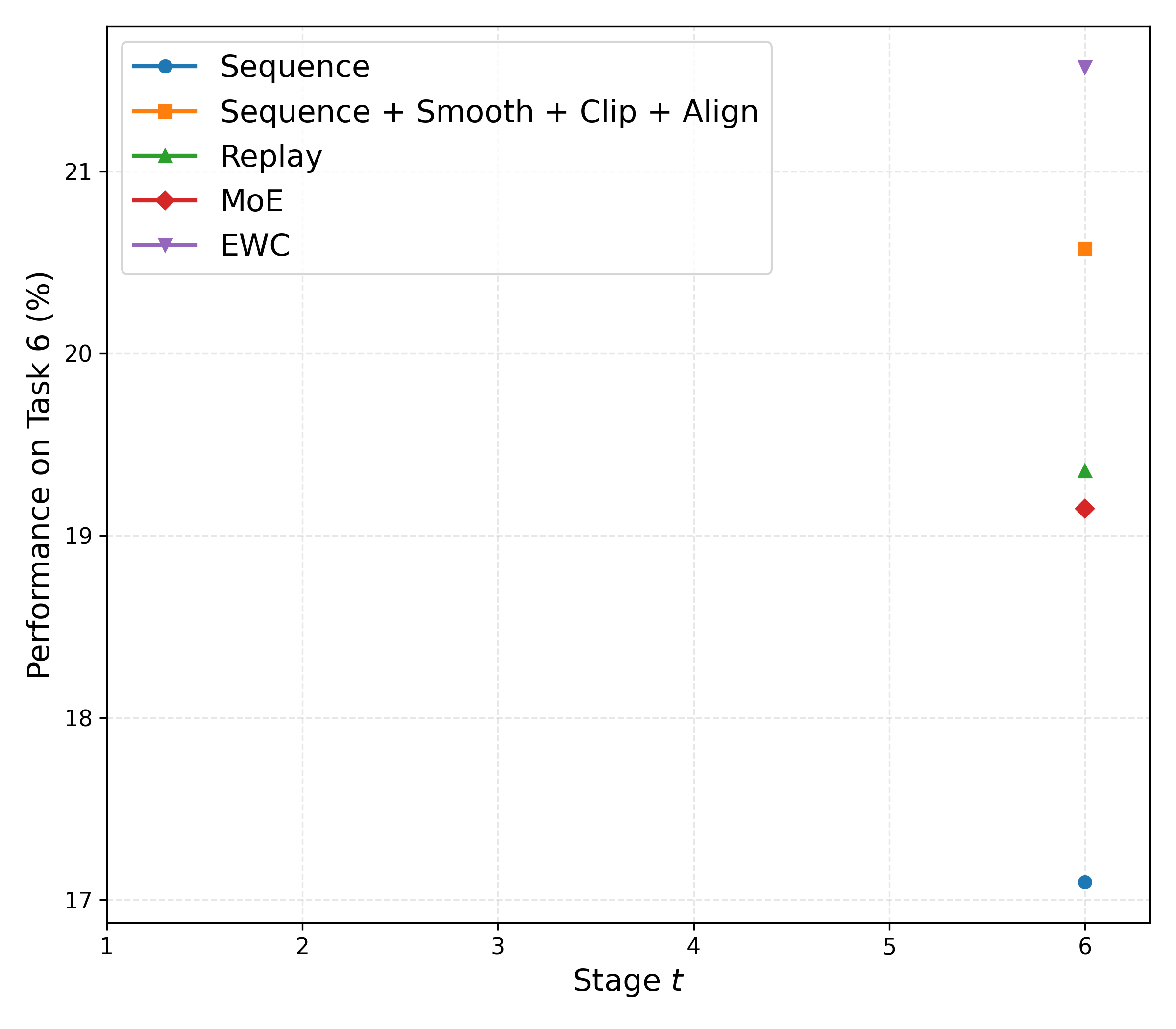}
    \caption{Task 6 (xlsum)}
    \label{fig:app_task6_curve}
  \end{subfigure}

  \caption{Per-task stagewise performance curves across all stages.}
  \label{fig:app_per_task_stagewise_curves}
\end{figure*}

Figure~\ref{fig:app_per_task_stagewise_curves} shows per-task stagewise performance curves. It can be observed that the three-module combination prioritizes long-term stability and cross-stage transfer rather than short-term optimality at each stage. Moreover, the improvements vary by task. On Tasks 1, 3, and 4, it yields more stable performance curves and maintains or even improves performance on earlier tasks in later stages, indicating stronger retention and potential positive backward transfer. For Task 5, while the performance is not particularly strong at the initial stage, it increases as stages progress. In contrast, the corresponding Sequence baseline shows a declining trend, suggesting that the benefits of our method may accumulate gradually over stages. Notably, degradation on Task 2 (emotion) is not unique to our method. Except for Replay, most baselines exhibit a similar declining trend on this task, which we attribute to this task’s fine-grained semantic boundaries and its dependence on lexical sentiment markers (e.g., negation, intensifiers, and affective words). These properties make it more vulnerable to representation drift caused by learning subsequent tasks. Meanwhile, Replay’s advantage is consistent with sample rehearsal, which explicitly reinforces earlier decision boundaries.

\subsection{Standard and Robust Performance Comparison}
\begin{figure*}[!t]
  \centering

  \begin{subfigure}[t]{0.7\textwidth}
    \centering
    \includegraphics[width=\linewidth]{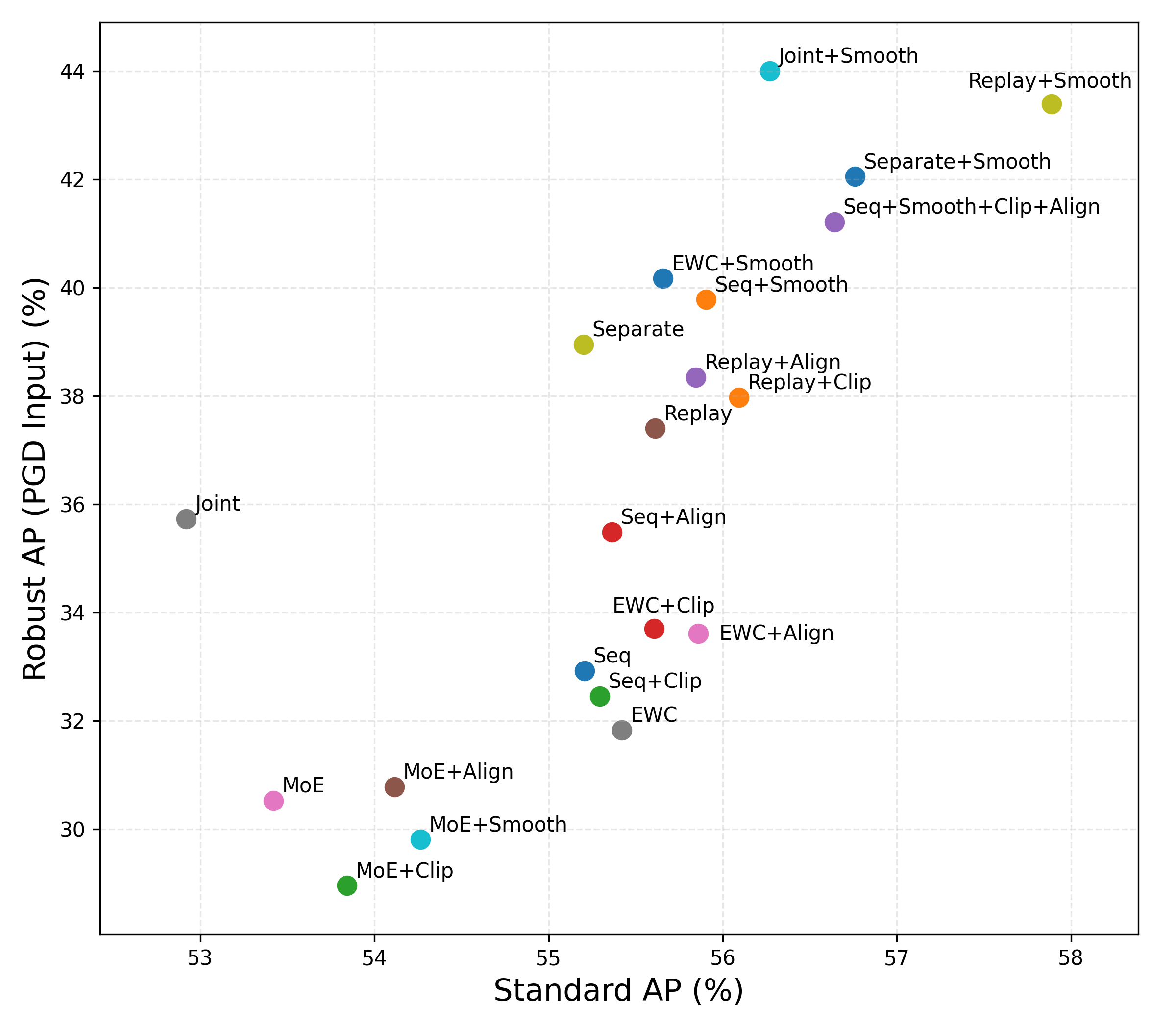}
    \caption{PGD (Input)}
    \label{fig:app_tradeoff_input}
  \end{subfigure}

  \vspace{0.8em}

  \begin{subfigure}[t]{0.7\textwidth}
    \centering
    \includegraphics[width=\linewidth]{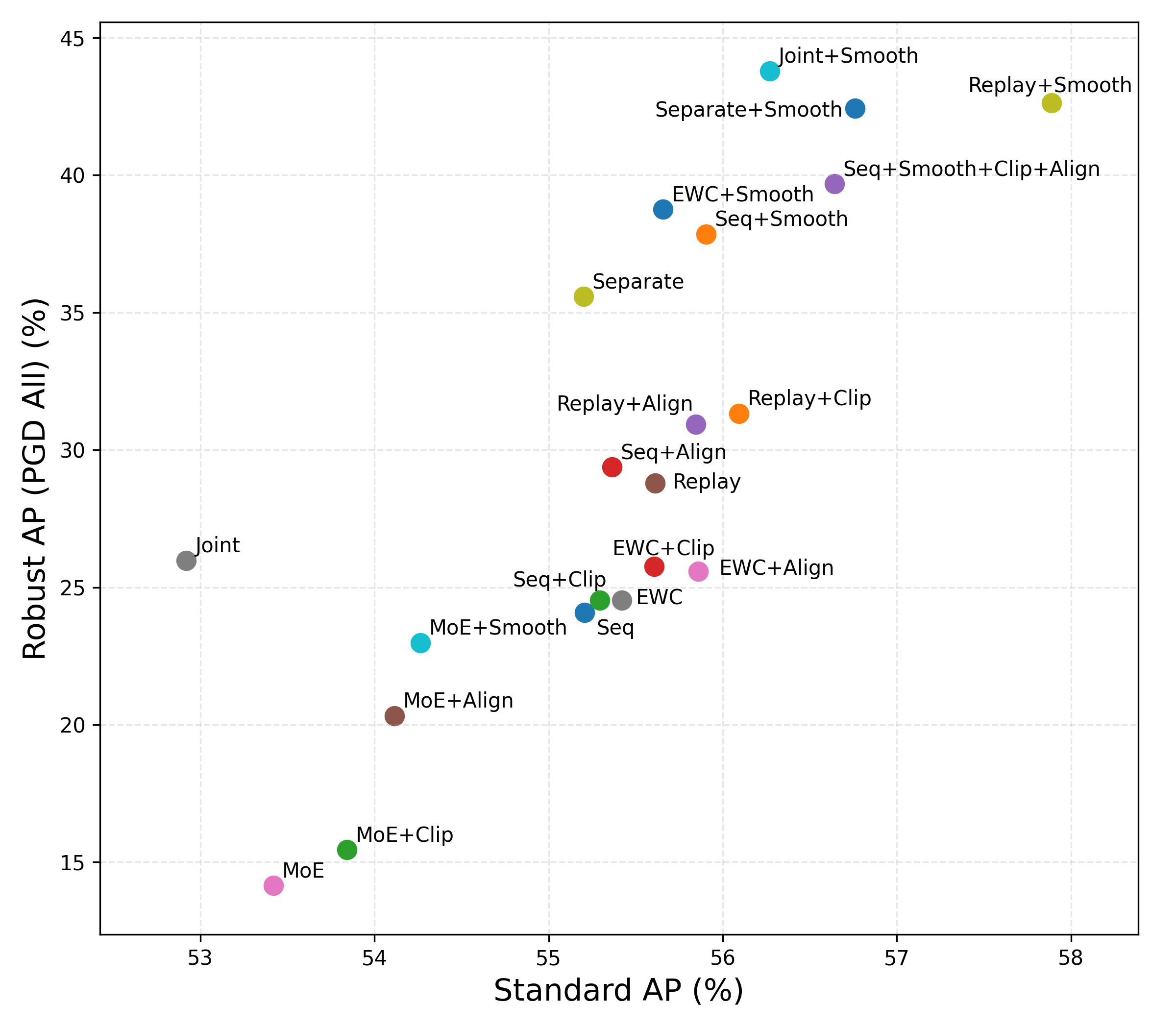}
    \caption{PGD (All)}
    \label{fig:app_tradeoff_all}
  \end{subfigure}

  \caption{Comparison between standard average performance and robust average performance under PGD attacks. PGD (Input) perturbs only input tokens, whereas PGD (All) perturbs both instruction and input tokens. Here, seq denotes the Sequence baseline.}
  \label{fig:app_tradeoff_clean_robust}
\end{figure*}

Figure~\ref{fig:app_tradeoff_clean_robust} presents a two-dimensional view of different methods by jointly comparing standard average performance (Standard AP) and robust average performance (Robust AP) under two robustness settings. Overall, the relationship between Standard AP and Robust AP is not a simple trade-off. In particular, most baselines enhanced by our modules move toward the upper-right region, reflecting simultaneous improvements in both Standard AP and Robust AP. Note that Intra-Smooth, Proto-Clip, and Inter-Align are not designed with identical goals. Proto-Clip acts as a regularizer that limits over-alignment to the current task, while Align targets cross-task directional alignment. Nevertheless, under robustness settings these modules can still affect Robust AP, which often shifts the enhanced baselines upward while maintaining competitive Standard AP. At the same time, only a few outliers are observed, mostly under PGD (Input) setting.

When the three modules are combined with Sequence baseline, the resulting model achieves higher Standard AP than the corresponding baseline and exhibits a consistent improvement in Robust AP. Among all the models, it is surpassed only by two special settings, Replay + Intra-Smooth, which benefits from rehearsal samples, and Separate + Intra-Smooth, which trains each task independently. The former directly reinforces earlier tasks via explicit replay, whereas the latter is not subject to cross-task forgetting, placing both at a more advantageous position in this comparison.

\end{document}